\pdfoutput=1

\documentclass[10pt, twocolumn, letterpaper]{article}

\usepackage{EMNLP2022}

\usepackage{times}
\usepackage{latexsym}
\usepackage{graphicx}
\usepackage{booktabs}
\usepackage[normalem]{ulem}
\usepackage{setspace}
\usepackage{stfloats}
\usepackage{nameref}
\usepackage{subfig}
\usepackage[symbol]{footmisc}

\renewcommand{\thefootnote}{\fnsymbol{footnote}}

\usepackage[T1]{fontenc}

\usepackage[utf8]{inputenc}

\usepackage{microtype}

\usepackage{amsmath}
\usepackage{longtable}
\usepackage{xcolor}
\usepackage{listings}
\usepackage{color}
\usepackage{soul}
\usepackage{multirow}
\usepackage{booktabs}
\usepackage{enumitem,kantlipsum}
\usepackage{amsfonts}
\usepackage{balance}
\usepackage{lipsum}

\newcommand\blfootnote[1]{%
  \begingroup
  \renewcommand\thefootnote{}\footnote{#1}%
  \addtocounter{footnote}{-1}%
  \endgroup
}

\definecolor{dkgreen}{rgb}{0,0.6,0}
\definecolor{gray}{rgb}{0.5,0.5,0.5}
\definecolor{mauve}{rgb}{0.58,0,0.82}
\definecolor{lightred}{RGB}{255,204,204}
\definecolor{lightgreen}{RGB}{204,255,204}
\definecolor{lightblue}{RGB}{204,255,255}

\DeclareRobustCommand{\hlred}[1]{{\sethlcolor{lightred}\hl{#1}}}
\DeclareRobustCommand{\hlblue}[1]{{\sethlcolor{lightblue}\hl{#1}}}

\sethlcolor{lightgreen}

\newcommand\thoughts[1]{}

\newcommand\amin[1]{\textcolor{blue}{#1}}
\newcommand\rindra[1]{\textcolor{cyan}{#1}}
\newcommand\tim[1]{\textcolor{violet}{#1}}

\newcommand\note[1]{\textcolor{red}{#1}}

\newcommand{\bftab}{\fontseries{b}\selectfont}

\title{Augmenting Operations Research with Auto-Formulation of \\ Optimization Models from Problem Descriptions}

\author{Rindranirina Ramamonjison$^1$\footnote[2]{} \And Haley Li\footnote[1]{} \And Timothy T. Yu\footnote[2]{} \AND Shiqi He\footnote[1]{} \And Vishnu Rengan\footnote[1]{} \And Amin Banitalebi-Dehkordi\footnote[2]{} \AND Zirui Zhou\footnote[2]{} \And Yong Zhang\footnote[2]{}}

\begin{document}
\maketitle

\blfootnote{\textsuperscript{1} rindranirina.ramamonjison@huawei.com}
\blfootnote{\textsuperscript{$\dagger$} Huawei Technologies Canada}
\blfootnote{\textsuperscript{*} University of British Columbia}
\blfootnote{\textsuperscript{\ddag} Code and data are available at the following \href{https://github.com/nl4opt/nl4opt-competition}{\textcolor{blue}{repository}}}

\begin{abstract}
We describe an augmented intelligence system for simplifying and enhancing the modeling experience for operations research. Using this system, the user receives a suggested formulation of an optimization problem based on its description. To facilitate this process, we build an intuitive user interface system that enables the users to validate and edit the suggestions. We investigate controlled generation techniques to obtain an automatic suggestion of formulation. Then, we evaluate their effectiveness with a newly created dataset\textsuperscript{$\ddag$} of linear programming problems drawn from various application domains.
\end{abstract}
\section{Introduction}\label{sec:intro}
Many real-world decision-making problems can be formulated and solved as mathematical optimization problems. The field of operations research (OR) has seen success in applications ranging from increasing bike-share ridership and efficiency \citep{bikeshare, bicycleGuangzhou}, managing wastewater collection and treatment systems \citep{wastewater}, to finding a revenue-maximizing pricing strategy \cite{revenue}. In fact, optimization solvers can tackle different types of problems as they are powered by efficient algorithms such as the simplex method \citep{simplex} or interior-point
methods \citep{karmarkar}.

\thoughts{\tim{modeling process and what is status quo and what problems we want to solve (Modeling is intensive). What do organizations need to do? First para of section 2.}
\tim{merging from below}}

However, modeling a problem into a proper formulation is a complex and time-consuming process. First, a domain expert must describe the problem and identify its variables, parameters, objective and constraints. \thoughts{Each element of the model such as the decision variable to optimize, the constraints and objective must be precisely communicated. \thoughts{Precisely, it must be clear what decision variables need to be optimized, what the constraints for the problem are, and what the objective is.}Additionally, the correct data parameters must be collected to quantify the dependencies of the objective and constraints with respect to the decision variables.} Then, an OR expert needs to translate this problem description into a precise formulation using a modeling language, thus making the process inefficient and limits the accessibility of the solvers to non-experts \cite{hurlimann2013mathematical}. \thoughts{Generally, an organization has a domain expert \thoughts{who is }responsible for describing the model elements and data parameters in the language of the problem domain. Then, the organization works with service vendors or OR consultants, who have a deep technical expertise in optimization techniques, to formulate the problem into a modeling language.} \thoughts{and can be used by commercial and\thoughts{or} open-source solvers}   \thoughts{makes this difficult.}

\thoughts{\tim{Here, we should present the augmented intelligence solution.}}

We propose an augmented intelligence system to simplify and enhance the modeling process. We partially automate the process using NLP models to suggest a formulation that the users can validate or edit the suggestions using an intuitive interface. Partial automation avoids the manual writing of the formulation by using a modeling language, thereby reducing the time and expertise to build optimization models. The intuitive interface also makes solvers more accessible to non-technical users. \thoughts{improve the performance and user experience in the model formulation process by bridging } \thoughts{allowing for manual model element revisions to correct or expand upon the predicted problem formulation.}\thoughts{We propose an augmented intelligence solution to improve the accessibility of solvers by bridging the language barrier between the domain expert and solvers.} \thoughts{Our system is purposefully not fully-automatic because of the challenges listed below that may cause the system to make mistakes. Hence, our}

\thoughts{\tim{Modeling is intensive. - can focus on augmented as a solution. Reason we want augmented solution is bc intensive - can focus on augmented as a solution. Reason we want augmented solution is bc intensive.  also mention highlight problems like AI can still make mistakes. Therefore human in loop.} \tim{Here, we can make it more concise. Do we have space for this? Have a more direct approach. say we propose an augmented solution. Focus of paper is to build an augmented modeling experience with a HITL system. What is the challenge for AI modeller? Need to pass problem description to formulation. Copy the goal. What the process works. We have tho hire people.. and waste time and money. Sales calls paper - 1st paragraph integral part is summarizing call summary. It's the number 1 time waster. So for us, having to formulate the problem and write in AML takes time and expertise.}}

From an NLP perspective, there are many challenges to parsing an optimization problem's formulation from its natural language description:

\begin{itemize}[wide, labelwidth=!, labelindent=0pt]
\item \textbf{Limited dataset.} The strenuous nature of modeling makes the cost of creating and labeling a large dataset prohibitive. Thus, 
efficient few-shot semantic parsers must be trained in a low-resource setting. \thoughts{Similar efforts and technical expertise are required to create a problem description and formulate the corresponding AML output. }Therefore, the solution must be built leveraging methods that excel on a limited training dataset.\thoughts{ methods are needed to build semantic parsers with a small training dataset. }

\item \textbf{Document-level input.} Most semantic parsers operate primarily at the sentence level.\thoughts{\rindra{citation needed}}
In contrast, long paragraph inputs describe the many variables, parameters, and constraints of optimization problems. Also, the parsing task involves a high level of compositionality and ambiguity. 

\item \textbf{Context-free output.}  \thoughts{\tim{move this up}}\thoughts{\note{a visual example would be nice here.}}\thoughts{Second, the output of the semantic parsing is an algebraic modeling language.}The outputs of most semantic parsing tasks share some contextual information with the inputs (e.g. database table or column names in SQL queries).  In contrast, our task has a context-free tabular format, which makes it difficult to align the input-output pair. 
\thoughts{\rindra{Clarify this format in Section 4: Precisely, it is a table where rows contain the parameters of the objective and constraints and each column header is a variable symbol or the right-hand-side limit of a constraint.}}

\item \textbf{Domain-agnostic parsing.} Finally, OR can tackle a diverse range of applications \cite{williams2013model}. Hence, the semantic parser must generalize well not only to new problem instances but also\thoughts{ to} new application domains. 
\end{itemize}

In this paper, we describe the underlying system that addresses these challenges and that enables the augmented modeling application. Our contributions are:

\begin{enumerate}[leftmargin=*]

\item A novel augmented intelligence application that simplifies and enhances the modeling process.
\item New controllable generation methods for parsing the formulation of optimization problems from its natural language description.
\item The first dataset on linear programming word problems, with which we test and analyze the effectiveness of the methods for this emerging application.


\end{enumerate}


\thoughts{\amin{Our modeling interface has been adopted in our commercial optimization solver,
which led to $\approx$30$\times$ modeling time improvements. To the best of our knowledge, this is the first time an AI-based modeling interface has been incorporated in an industrial solver.}}\thoughts{\tim{If same text in abstract, change this up a little bit}}

\begin{figure*}
\centering\includegraphics[width=1.75\columnwidth]{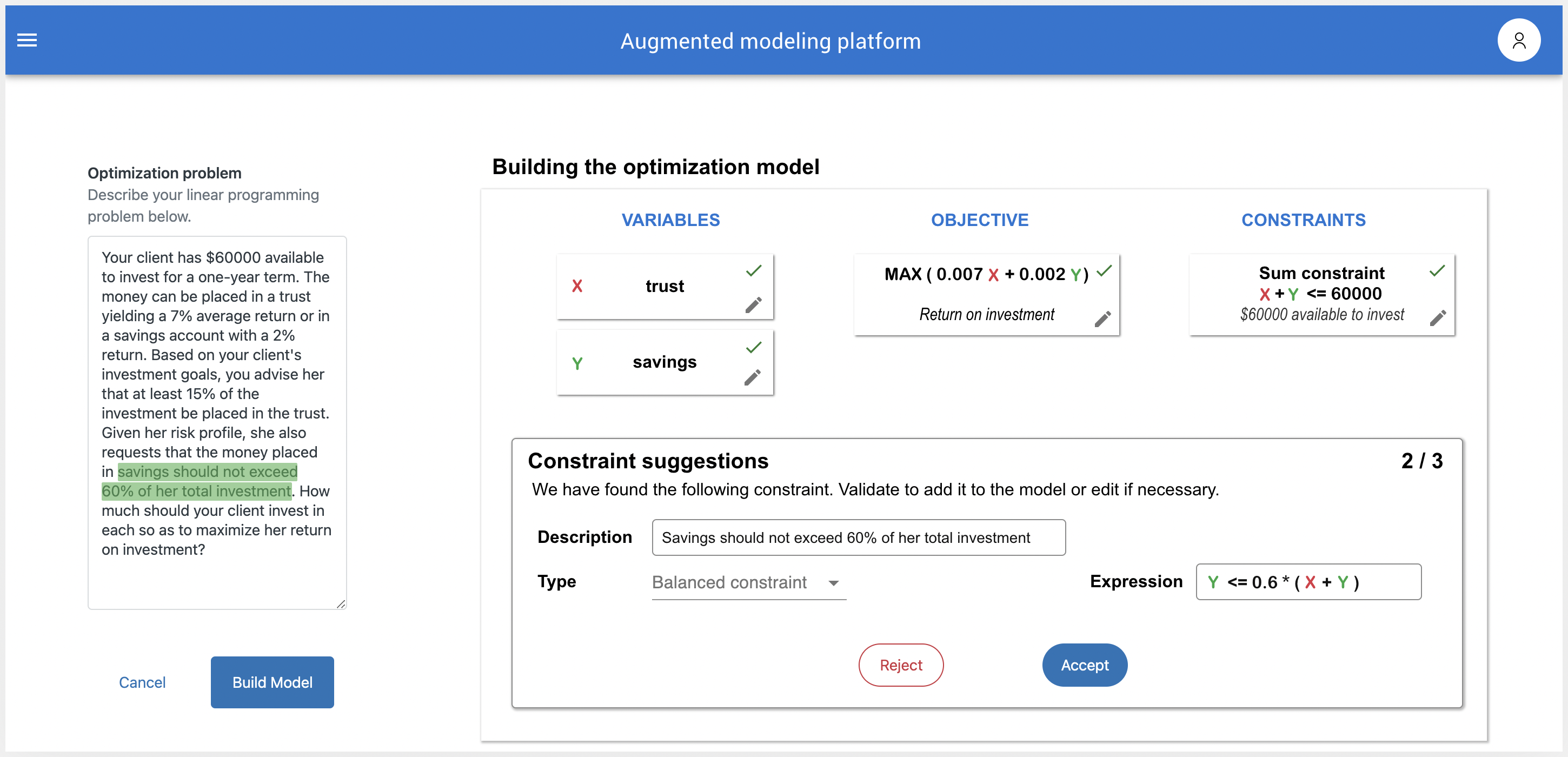}
\vspace{-10pt}
\caption{Augmented modeling platform.}
\label{fig:app}
\end{figure*}

\section{Related Work}
\thoughts{\tim{perhaps cut down on related work to expand more on the industry impact or to have more space for the dataset? - but reviewer liked why it's still different from traditional existing semantic parsing one.} \tim{check .md file - reduce repetition between sections}}

\paragraph{Augmented intelligence systems.}
Augmented intelligence systems use AI to assist (and not replace) the users in performing certain tasks. These systems could improve the experience or creativity of users in artistic applications such as story writing \cite{Clark2018CreativeWW}, music composition \cite{https://doi.org/10.48550/arxiv.2010.05388}, poetry composition \cite{uthus-etal-2022-augmenting}, and sketching \cite{10.1145/3325480.3326578}. This approach has also seen success in more task-oriented applications by helping teachers to grade homework efficiently \cite{DBLP:conf/edm/MalikWVSCMGP21}, salespeople to summarize sales calls \cite{asi-etal-2022-end}, and improving pneumonia diagnostic accuracy\thoughts{ compared to the current gold standard } \cite{osti_10125837}. We adopt a similar approach for OR and focus on improving the modeling process. \thoughts{Through this framework, we improve the performance of the generated model suggestion when compared to an AI-only approach.}


\paragraph{Semantic parsing and generation.}
Semantic parsing \thoughts{plays an important role in building natural language interfaces for different systems\thoughts{. In fact, its goal is to map} by mapping }maps natural language utterances into a \thoughts{structured and }machine-interpretable representation \cite{kamath2019survey}. This mapping has been extensively studied for output representations such as SQL queries  \cite{gan-etal-2020-review}\thoughts{\cite{zhong2017seq2sql-custom, gan-etal-2020-review}}, Unix commands \cite{bharadwaj-shevade-2021-explainable}\thoughts{\cite{lin-etal-2018-nl2bash, bharadwaj-shevade-2021-explainable}}, or logical forms for querying a knowledge base \cite{dong-lapata-2016-language}. We tackle a different and challenging semantic parsing task as explained in Section \ref{sec:intro}.

Building on the success of attention models in sequence-to-sequence tasks \cite{sutskever2014-custom, luong-etal-2015-effective}\thoughts{\cite{sutskever2014-custom, bahdanau2015-custom, luong-etal-2015-effective}}, encoder-decoder architectures have been adopted for designing semantic parsers \cite{dong-lapata-2016-language, dong-lapata-2018-coarse, wang-etal-2020-rat}\thoughts{\cite{dong-lapata-2016-language, jia-liang-2016-data, dong-lapata-2018-coarse}}. \thoughts{Furthermore, the use of large pre-trained language models \cite{devlin-etal-2019-bert, xue-etal-2021-mt5}\thoughts{\cite{devlin-etal-2019-bert, lewis-etal-2020-bart, gpt3-custom, xue-etal-2021-mt5}} have significantly impacted the performance of semantic parsing. For example, \cite{wang-etal-2020-rat} built an encoder-decoder model using the BERT transformer and \cite{scholak-etal-2021-duorat}\thoughts{\cite{scholak-etal-2021-duorat, rongali2020-custom}} enhanced the decoder using pointer networks \cite{vinyals2017pointer-custom}\thoughts{\cite{vinyals2017pointer-custom, see-etal-2017-get}}.} We propose \thoughts{controlled generation methods that perform the mapping }using an intermediate representation (IR) \thoughts{and}that serves as a bridge between natural language and the canonical output format. Our two-stage mapping strategy is different from \cite{dong-lapata-2018-coarse}, which initially generates a sketch of the query and then fills out the slots. In contrast to prior methods of constrained decoding \cite{hokamp-liu-2017-lexically, scholak2021picard-custom}, our approach uses a simple beam search and leverages a prompt-guided generation and copying mechanism to guide the decoding\thoughts{ and improve the accuracy of the mapping}. \thoughts{We construct the prompt as a prefix of target IR based on the tagged entities of the problem description which can be extracted via a HITL framework\thoughts{, which can be provided by an entity recognition model or by the user via an user interface}.}

\paragraph{Datasets on Mathematical World Problems (MWP).}
Recent works have studied the use of NLP models to automatically solve MWP \cite{wang-etal-2017-deep, surveyMWP-custom} Most existing \thoughts{datasets on MWP}MWP datasets have focused on returning the solutions of elementary arithmetic problems \cite{roy-roth-2015-solving, koncel-kedziorski-etal-2016-mawps}\thoughts{\cite{roy-roth-2015-solving, hosseini-etal-2014-learning, koncel-kedziorski-etal-2016-mawps}} and algebra problems \cite{kushman-etal-2014-learning, huang-etal-2016-well}. More challenging benchmarks have been recently proposed such as SVAMP \cite{patel-etal-2021-nlp}, MATH \cite{hendrycksmath2021-custom} and GSM8K \cite{trainingVerifiers2021-custom}. In contrast, we build the first linear programming word problems (LPWP) dataset and evaluate methods\thoughts{ to gen algebraic formulation of an optimization problem \thoughts{because we want }} of generating the formulations as inputs to optimization solvers which can efficiently return an optimal solution.\thoughts{to optimization solvers, which can efficiently return an optimal solution using scalable algorithms.}\thoughts{ To this end, we build the first dataset of linear programming word problems, or LPWP in short, for demonstrating. We build the first LPWP dataset to investigate this emerging NLP application. }\thoughts{In contrast to \cite{drori2021neural-custom}, we want to generate a set of declarations instead of an imperative program composed of a sequence of operations. }

\begin{figure*}
\centering\includegraphics[width=2\columnwidth]{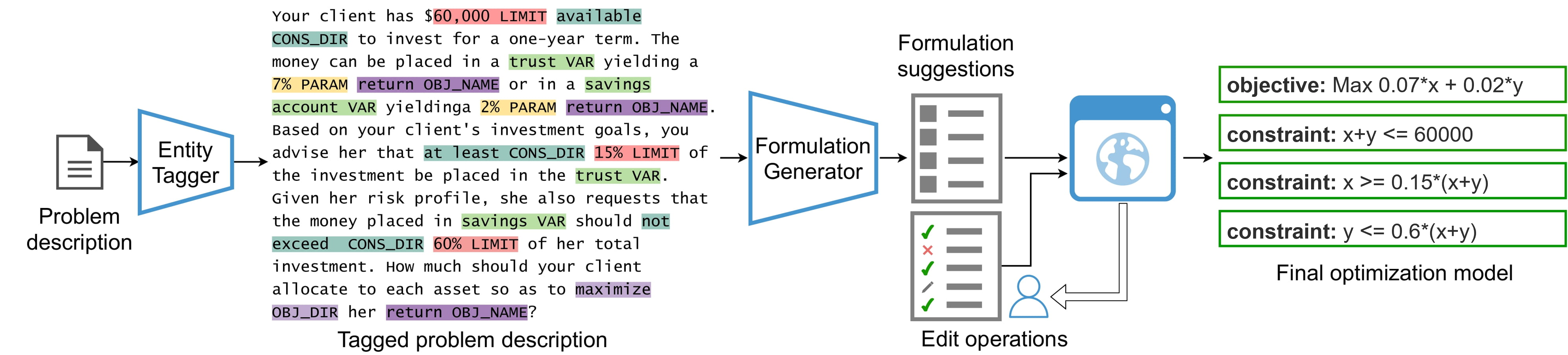}
\vspace{-10pt}
\caption{System diagram for augmented modeling framework.}
\label{fig:system}
\end{figure*}

\section{Augmented Modeling Interface}

We present an interactive system that enables users to model an optimization problem in collaboration with an AI system. To use this application, the user first describes the problem using natural language.\thoughts{\tim{is this slow? will there be big problems that this one-by-one approach will be tedious?}} Then, the system will suggest to the user the formulation of the optimization model including the decision variables, objective, and problem constraints. The system suggests the components of the formulation one at a time allowing the users to accept, reject, or edit the suggestions. 

An example scenario of a portfolio optimization problem is shown in Figure \ref{fig:app}. Here, the system suggested the formulation of a balance constraint given the description ``\emph{savings should not exceed 60\% of her total investment}''. Had the system presented an incorrect formulation (e.g. an upper bound constraint), the user can change the type using the interface and the system reformulates the constraint expression automatically. The user can also edit the description, which will be stored as metadata of the model. In the same fashion, the user can add variables or constraints manually or edit the problem description to update the model. When the user is satisfied with the formulation, the system forwards it to an optimization solver, which then returns either an optimal solution of the problem to the user or warnings for some infeasible constraints.

Figure \ref{fig:system} shows an overview of the underlying auto-formulation system. It consists of an entity tagger, a formulation generator, and an augmented modeling interface. Given a problem description, the entity tagger labels the keywords that indicate the components of the optimization problem. For example, ``return'' and ``at least'' are tagged as an objective name and constraint direction, respectively. Then, the formulation generator uses the text description and the corresponding tagged entities to generate the formulation suggestions, which are then presented to the user by the augmented modeling interface. In our implementation (experimental settings in the \nameref{sec:appendix}), we used an XLM-RoBERTa pre-trained transformer and fine-tuned it for entity recognition using the dataset described in Section \ref{sec:results}. \thoughts{\rindra{Do we need to specify NER model implementation and its F1 score here: it achieved a micro-averaged F1 test score of 84\%.}}\thoughts{\tim{Might want to cut this since it's already introduced in the next section} In the next section, we describe in detail the different approaches and methods used to implement the formulation generator.}

\begin{figure*}
\centering\includegraphics[width=2\columnwidth]{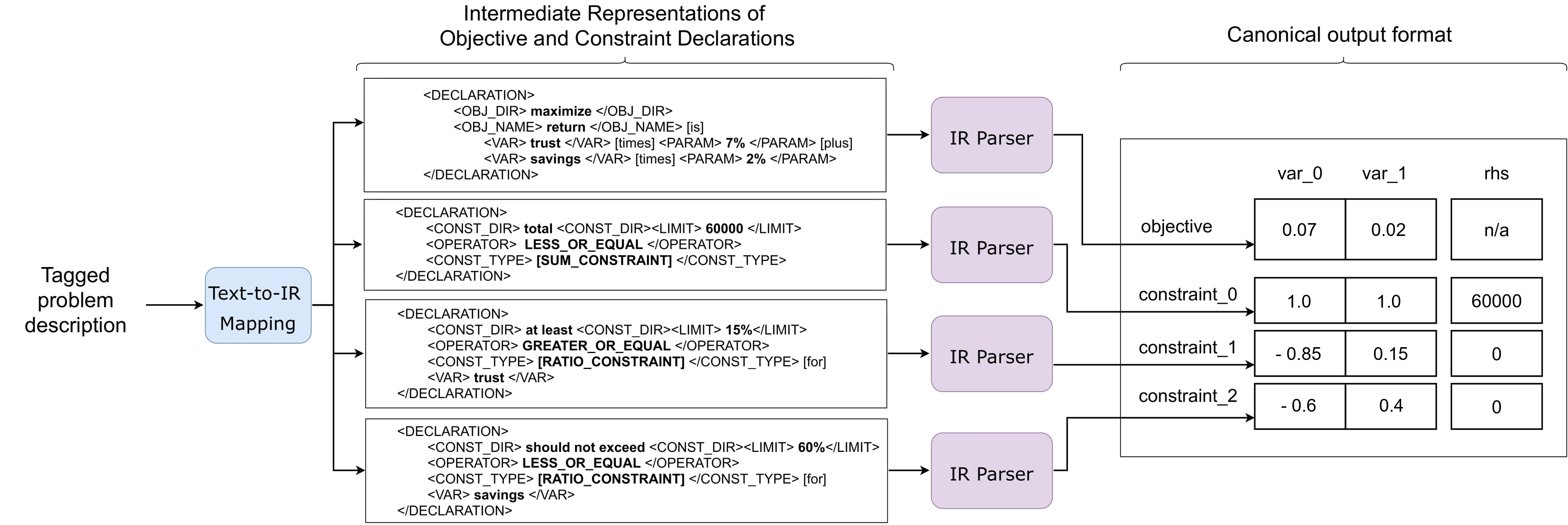} 
\vspace{-5pt}
\caption{Overview of our formulation generation approach.}
\label{fig:overview-generation}
\end{figure*}

\thoughts{\rindra{The same LPWP dataset, as described in Section \ref{sec:dataset-creation}, was tagged for its problem entities and used to train a named entity recognition (NER) model. The NER model leveraged a XLM-RoBERTa pre-trained transformer. When evaluated on the reserved test split, it achieved a micro-averaged F1 score of 0.905 and 0.815 on the in-domain (source domain) and out-of-domain (target domain) splits, respectively. Additional details can be found in the \nameref{sec:appendix}}}

\thoughts{\rindra{We propose a modeling interface where a user can interact by entering a problem description and the system transform it into an equivalent formulation. For this to work, it requires a prompt for decoding the declarations. To this end, we trained a named entity recognition model to recognize the optimization problem entities. Then, we use the later to infer the declaration prompts. In addition, the users may also be given an interface where they can tag the entities of the problem to refine the prompt creation. For future work, we will also consider implementing a validation system that detects errors in the generated formulation and highlights the invalid tokens. A complete example of an interactive workflow is illustrated in the \nameref{sec:appendix}.} \thoughts{\rindra{Add figure in Appendix or supplementary result}}}

\section{OptGen: Controllable Generation of Optimization Formulation}\label{sec:method}

Here, we present the methods behind our model OptGen, for generating the suggested formulation.

\subsection{Two-stage mapping approach}
It is difficult to directly map the problem description $p$ to the formulation $f$ due to the characteristics of the  input-output pair $\left(p,f\right)$ as mentioned in Section \ref{sec:intro}. First, the input document $p$ can be unstructured and ambiguous especially when it describes many constraints of different types. From the input, we must precisely extract the canonical representation $f$ of an optimization formulation. 

As shown on the right of Figure \ref{fig:overview-generation}, this canonical representation is a context-free table in which the column header is either a variable symbol or a constraint's right-hand-side (rhs) limit. Each table row contains the parameters of the objective function or constraint. As a result, the canonical formulation $f$ is context-free since it abstracts away the contextual information of \thoughts{the problem description} $p$.

Instead, we adopt a two-stage mapping $p\mapsto r \mapsto f$ to bridge the gap between the natural language input and context-free formulation. 
\paragraph{Text-to-IR mapping.} We first define an intermediate representation (IR) $r$ of the problem to simplify the parsing. As illustrated in Figure \ref{fig:overview-generation}, a Text-to-IR mapping model generates a set of entity-typed declarations $\left\{ D_{i}\right\} _{i=1}^{n}$ defined in an extended markup format to simplify its parsing. Note that other formats can be used for the IR (e.g. a format defined by a context-free grammar). \thoughts{\rindra{citation}}Each declaration $D_{i}$ is a sequence of tokens that represents a typed and structured representation of either the optimization objective or a constraint. Each $D_{i}$ is defined based on a predefined template of the different objective and constraint types.  
The tokens defining each declaration are wrapped in special tags. For example,\thoughts{of tokens} \thoughts{the sequences} \lstinline{<OBJ_DIR> maximize </OBJ_DIR>} and \lstinline{<OBJ_NAME> return </OBJ_NAME>} define the direction and the name of the objective. 
Since these tokens are derived from the input description, the IR preserves the\thoughts{contextual information} context of the problem. We describe the grammar of the IR in the \nameref{sec:appendix}. 

\paragraph{IR parsing.} An IR parser converts each IR declaration to the canonical format. We use an XML parser and apply simple transformations to convert numerical words to decimal numbers and to follow the following conventions. First, the canonical format always "minimize" a cost function. When the objective direction is "maximize," we instead change the sign of each objective parameter. Similarly, each inequality constraint must have a \lstinline{LESS_OR_EQUAL} operator and convert each inequality constraint to the form $a^{\top}x \le b$.

\subsection{Prompt-guided generation model}
\begin{figure*}
\centering\includegraphics[width=2\columnwidth]{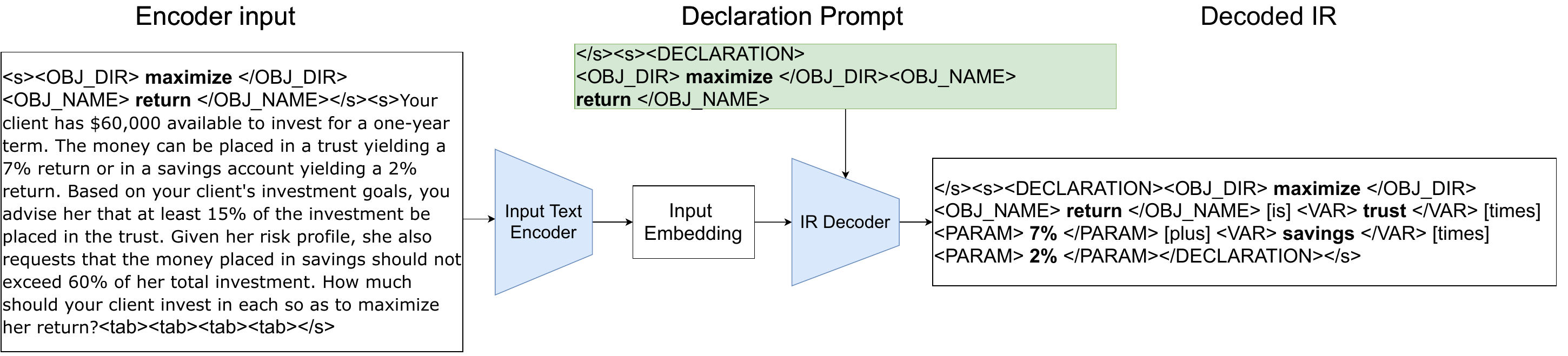} 
\vspace{-5pt}
\caption{Description of prompt-guided generation method.}
\label{fig:prompt_guided}
\end{figure*}

We use an autoregressive model that is built upon the BART language model \cite{lewis-etal-2020-bart}. A prompt-guided generation is proposed to improve the accuracy of the Text-to-IR mapping. The idea is to decode the declarations of the IR one by one by using a declaration prompt to focus the generation. A declaration prompt is a prefix of the IR declaration of an objective or a constraint. For an objective, the prompt is composed of the entity tokens of the objective's direction and name. Similarly, the prompt for a constraint is defined by the entity tokens of the constraint direction. These tokens are obtained from the output of the Entity Tagger model shown in Figure \ref{fig:system}. The declaration prompt is added to the input text of the encoder. The role of the prompt is to provide contextual triggers for the decoder to focus on the relevant parts of the declaration to be generated. As illustrated in Figure \ref{fig:prompt_guided}, the model is trained to generate the IR of the declaration based on the declaration prompt and the problem description.

One key requirement of the Text-to-IR mapping is the ability to extract the variable names and data parameters from the descriptions and copy these important mentions from the input description into the output IR of the decoder. To augment the capability of BART encoder-decoder model, we leverage a copy mechanism that computes the probability distribution $P_{\text{copy}}$ over the input tokens using cross-attention scores. The copy distribution is calculated at each time step $t$ by taking the mean of the decoder’s cross-attention scores across all attention heads as follows:
\vspace{-5pt}
\begin{align*}
e_{t,i}=\frac{\left(W_{s}s_{t}\right)^{T}W_{h}h_{i}}{\sqrt{d_{k}}}\\
\alpha_{t,i}=\text{softmax}\left(e_{t,i}\right)\\
P_{\text{copy}}=\frac{1}{n_{H}}\sum_{i}\alpha_{t,i}
\end{align*}
where $W_{s}$ and $W_{h}$ are the projection matrices for the encoder
and decoder. Then, $s_{t}$, $h_{i}$, and $n_{H}$ are the decoder
hidden state at time step $t$, the encoder hidden state for the attention
head $i$, and the number of heads respectively. 

We add the special tokens of the IR into the BART target vocabulary and mask out any vocabulary words that are not present in the source input. Following \cite{see-etal-2017-get}, we use a soft switch $p_\text{gen}\in\left[0,1\right]$ to choose between generating a word from the vocabulary by sampling from $P_\text{vocab}$, or copying a word from the input sequence by sampling from $P_\text{copy}$. Thus, the final probability distribution of a word $w$ is given by:
\vspace{-5pt}
\begin{equation*}
\hspace{-4pt}P\left(w\right)=p_{\text{gen}}P_{\text{vocab}}\left(w\right)+\left(1-p_{\text{gen}}\right)P_{\text{copy}}\left(w\right)
\end{equation*}
and is used to compute the loss for timestep $t$ as the negative loglikehood of the target word $y_t$ for that timestep during training. Then, we average the loss over all time steps. As a result, the model is trained to produce tokens from either the IR vocabulary or the input problem description.

\section{Experiments}
\subsection{Dataset}\label{sec:dataset}
\thoughts{\tim{NOT a section but subsection of experiments. - describes dataset, creation, and any screenshots to show is in appendix E has hyperparameters}}


\paragraph{Dataset description.}We curated a first-ever LPWP dataset\footnote{We have released this dataset and code to the research community for future research.} and use it to train and evaluate our methods. \thoughts{Don't think we have to specify this part: and also will use to launch an ML competition on the proposed task in the near future.}\thoughts{\paragraph{Description:}}The dataset contains a collection of LP problems of the form:
\vspace{-3pt}
\begin{align*}
    \min_{\mathbf{x}\in\mathbb{R}^{n}}     \mathbf{c}^{\top}\mathbf{x} \enskip\text{s.t.}\enskip
    \mathbf{a}_{i}^{\top}\mathbf{x}\leq b_{i},\enskip i=1,\ldots,m
\end{align*}

\noindent
where $\mathbf{c}$ and $\mathbf{a}_{i}$ represent the parameters of the objective and the $i$-th constraint, respectively. $b_i$ is the right-hand-side (rhs) limit, and the goal is to find\thoughts{the vector} $\mathbf{x}$ that minimizes the objective value. The objective and constraint functions are linear with respect to variables in the LP problems. Each example has a text description of the problem and is annotated with the IR, math representation, and canonical formulation as shown in Figure \ref{fig:overview-generation}. Table \ref{tab:statistics} summarizes the statistics of the dataset. More details and examples of the dataset are provided in the \nameref{sec:appendix}.

The dataset contains 1101 LP problems from the source domain (advertising, investment, sales) and target domain (production, science, transportation). The train, dev, and test splits contain 713, 99, and 289 samples, respectively. The training split is comprised of solely of samples from the source domain whereas the dev and test splits contain samples from both source and target domains with a source-to-target domain ratio of 1:3. 

\begin{table}[h!]
\centering
\fontsize{9.5}{10}\selectfont
\begin{tabular}{lc}
\toprule
Number of Problems      & 1101   \\ 
Number of Declarations        & 4216  \\ 
Number of Constraint Types     & 6     \\ 
Average Number of Variables   & 2.08 \\  
Average Number of Constraints & 2.83 \\
\bottomrule
\end{tabular}
\vspace{-8pt}
\caption{Summary statistics of the LPWP dataset.}
\label{tab:statistics}
\end{table}
\thoughts{\rindra{
\begin{enumerate}
    \item Describe and explain how we have created the dataset and what quality control we have taken to make sure we have some desired characteristics of the dataset (e.g. diversity of problem structures, etc.)
\end{enumerate}
}}

\begin{figure*}
\centering\includegraphics[width=1.70\columnwidth]{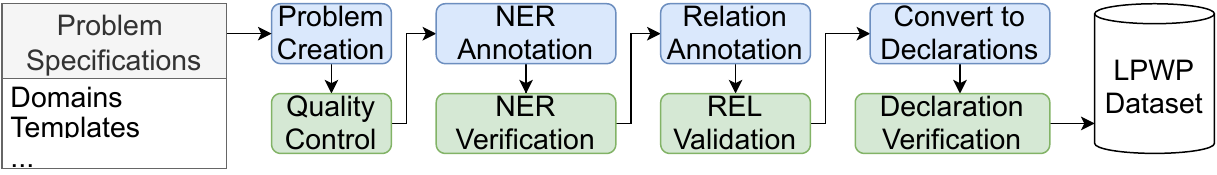}
\vspace{-5pt}
\caption{Overview of dataset creation.}
\label{fig:dataset-creation}
\vspace{-3pt}
\end{figure*}

\paragraph{Dataset creation and quality control.}\label{sec:dataset-creation}\thoughts{don't say exact but mention that IAA and other things are in the appendix}The dataset was created from scratch and annotated internally by a team of $20$ researchers and engineers with different levels of expertise in OR/NLP. \thoughts{Note that the dataset does not contain any personal information. }The process included 3 stages: (1) problem creation, (2) NER annotation, and (3) REL annotation and declaration generation. Each stage was followed by rigorous quality checks and verification. The creation process is described in Figure \ref{fig:dataset-creation}. 

Additional details of the dataset including the creation process, examples of the problem and their corresponding math formulation and IR, exclusion criteria, and inter-annotator agreement score, etc, are reported in the \nameref{sec:appendix}.


\subsection{Results and Discussions}\label{sec:results}
\paragraph{Baseline and metrics.} We conducted experiments using the dataset described in Section \ref{sec:dataset}. We use BART model \cite{lewis-etal-2020-bart} as baseline for the two-stage mapping approach. In addition, we adopt the Text-to-Table (T2T) model \cite{wuetal2021} as a baseline for the direct approach that directly produces the canonical form. An example of its output is shown in Table~\ref{appendix:table:texttotable} in the \nameref{sec:appendix}.

For evaluation, we measure the declaration-level mapping accuracy on the canonical formulation defined as:
\vspace{-5pt}
\[ \text{Acc} = 1 - \frac{\sum_{i=1}^N \min\left\{\text{FP}_i + \text{FN}_i, D_i\right\}}{\sum_{i=1}^N D_i},\]
where for a given problem $i$, $D_i$ is the number of ground-truth declarations, false positives $\text{FP}_i$ is the number of non-matched predicted declarations, and false negatives $\text{FN}_i$ is the number of excess unmatched ground-truth declarations. In other words, $\text{FN}_i$ is only non-zero when there are more ground-truth declarations than predicted declarations. The $\min$ is to prevent negative accuracy and overpenalization on single problems.

\paragraph{Main Results.} Table~\ref{tab:performance} summarizes our results on the source and target domains. The BART baseline achieved the worst performance on all domains. In fact, BART hallucinated by producing too many constraints or many wrong parameters. Next, the direct mapping approach of T2T model achieved the highest accuracy of 88\% on the Source domain but generalized poorly on out-of-domain (Target) test set. While it seemed to learn the task, T2T overfit to the Source training data. In contrast, our proposed OptGen achieved an absolute +18\% accuracy improvement over direct mapping (T2T) on the Target domain. It was able to generalize better than the other models and produced the fewest errors on the problem structure and parameters. Our two-stage approach helped in this regard while the T2T must directly learn to convert to the canonical form. A detailed error analysis can be found in Figures~\ref{appendix:fig:problemlevelerrors} and \ref{appendix:fig:declarationlevelerrors} in the \nameref{sec:appendix}.

\paragraph{Ablation study.} We also analyzed the importance of the individual methods used in our model in Table~\ref{tab:performance}. It shows that the copy mechanism is important for accurately mapping the description to the equivalent formulation. Without the copy mechanism, we see significant accuracy drops of about 5\% and 30\% on Source and Target respectively. We show a qualitative comparison of generated IR formulations for two LP problems in Tables\thoughts{the figures} \ref{appendix:table:furnituretable} and \ref{appendix:table:flowertable} in the \nameref{sec:appendix}. While our model could perfectly generate the correct representations, the model without copy mechanisms produced many errors. For example, it hallucinated the wrong constraint limits or detected the wrong constraint types. The prompt-guided generation method also led to slightly better performance on the Target domain. Finally, we noticed +6\% improvement for T2T on the target domain when using contextual prompts.

These results show the importance of using controlled generation techniques for learning the syntax and grammar of the target IR language and for accurately mapping the input description to the IR formulation. 



\begin{table}
\centering
\fontsize{8}{10}\selectfont
\begin{tabular}[t]{lrrrrr}
\toprule
Method & Source & Target & Sci & Prod & Trans\\
\midrule
T2T & 0.83 & 0.39 & 0.46 & 0.37 & 0.37 \\
T2T + Prompt & \bftab 0.88 & 0.45 & 0.49 & 0.46 & 0.40 \\
BART & 0.52 & 0.20 & 0.21 & 0.19 & 0.20\\
OptGen w/o copy & 0.55 & 0.34 & 0.38 & 0.32 & 0.33\\
OptGen w/o prompt & 0.58 & 0.61 & \bftab 0.64 & 0.63 & 0.57\\
OptGen & 0.60 & \bftab 0.63 & 0.60 & \bftab 0.66 & \bftab 0.64 \\
\bottomrule
\end{tabular}
\vspace{-8pt}
\caption{\label{tab:performance}\small Results for each model on the declaration-level mapping accuracy metric. Source consists of samples from the source domain test split. Target \thoughts{consists of samples from the target domain test split and }is a weighted mean (by number of declarations) of the science (Sci), production (Prod), and transportation (Trans) target domains test split.}
\vspace{-8pt}
\end{table}

\paragraph{Limitations and future works.} Our preliminary results show the potential and the importance of controllable generation methods for enabling the augmented modeling system. While the proposed OptGen generation model was shown to generalize better than baseline models, its accuracy performance should still be enhanced further by improving the decoding method or by using edit-based models \cite{https://doi.org/10.48550/arxiv.2206.07043} to automatically correct the erroneous parts of the formulation. As future work, we will conduct human evaluation of the augmented system by measuring the efficiency improvement perceived by real users. As the current dataset only covers LP problems, we will also expand it to cover other types of problems such as mixed-integer programs, which have different types of constraints. Other directions can also be explored to build more data-efficient methods.
\thoughts{\note{This paragraph can also go to the supplementary if there are space limitations. Just be careful to put the correct address in the responsible-NLP PDF file.}}

\section{Conclusion}
We introduced an augmented modeling platform, in which users enter the descriptions of optimization problems and interact with an AI system to efficiently model their formulations. To this end, we described the underlying system for this emerging application and proposed controllable generation methods to enable it. We also created a training dataset of linear programming word problems to evaluate the effectiveness of the proposed methods. Our findings showed that the design of generation models and methods can have significant impact on the accuracy of the system's suggested formulations and that the system should help the users to validate and edit the suggestions. 

\section{Ethics Statement}
This augmented intelligence system is intended to parse the formulation of optimization problems from its natural language description to aid stakeholders in their decision-making. The dataset was created taking special care to exclude samples with inappropriate language or names of real people, products or companies. The harm to users resulting from incorrect parsing is limited. However, depending on the application, the system may be used in sensitive or critical applications, such as a power grid, flights scheduling, etc. In such cases, the solver should be used with caution and the modeling process should be validated by the domain expert. Finally, operations research has historically been applied in tactical military operations. We must understand the potential negative impact of misusing this technology for society at large and the users must seriously consider the ethical concerns related to military applications. 



\bibliographystyle{acl_natlib}
\bibliography{anthology.bib, custom.bib}

\clearpage

\section{Appendix} \label{sec:appendix}
This section contains supplementary materials including qualitative comparisons, examples from the dataset, as well as additional details which were omitted from the main paper due to space limitations.

\appendix


\section{More details on dataset description}
In Table \ref{constraint-table}, we showcase the different constraint types, along with their use in a problem, their math formulation, and their IR. Tables \ref{appendix:table:resourcedomain}, \ref{appendix:table:investmentdomain} and \ref{appendix:table:farmingdomain} provide examples showing the problem description, IR and the math formulations.

\section{More details on dataset creation}
The LPWP dataset creation process consists of three stages.

\subsection{Stage 1: Problem creation}
First, we invited 15 researchers and engineers to create original LP problems. They created the problems following some specifications such as problem domains and templates. The problem domains include advertising, investment, sales, agriculture, manufacturing, transportation and health sciences. During the creation step, 5 additional people were tasked with performing quality control and giving guidance to the problem creators. This included screening for the exclusion criteria (e.g. use of inappropriate language or names of real people, products or companies) and making sure the produced problem followed the specified domain and template. In fact, we used these templates to ensure the diversity of the problem structure (e.g. number of constraints, objectives and constraint types) in the dataset. If a mistake was detected in the problem, the problem creator would have been asked to fix it and the correction was verified. We made sure that each problem was verified by at least two people. 

\subsection{Stage 2: NER Annotation}
\begin{figure*}[h!]
\centering\includegraphics[width=2\columnwidth]{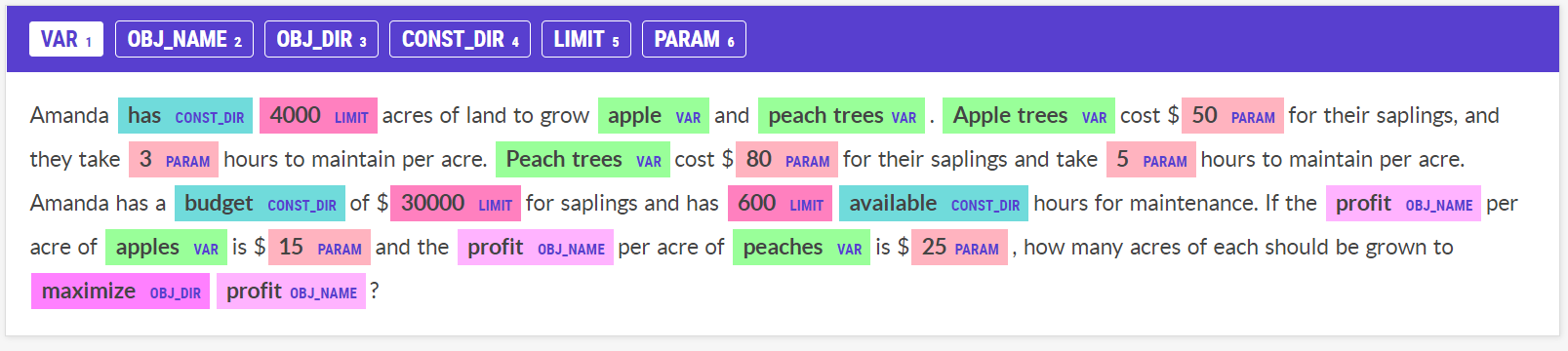}
\caption{Screenshot of NER annotation using Prodigy.}
\label{appendix:fig:ner-annotation-screenshot}
\end{figure*}

At this stage, annotators were required to locate and classify entities mentioned in the problem. There are in total 6 types of targeted entities: variable (VAR), parameter (PARAM), limit (LIMIT),  constraint direction (CONST\_DIR), objective direction (OBJ\_DIR) and objective name (OBJ\_NAME). We used the Prodigy annotation tool to annotate the problems. A screenshot of one annotation session is provided in Figure \ref{appendix:fig:ner-annotation-screenshot}. We asked an additional annotator to resolve disagreements by reading the guideline thoughtfully and asking the annotators for clarifications if needed.

To ensure the quality of the NER annotations, four OR/NLP experts annotated more than 10\% of the entire dataset, with an equal split between each domain, separately to compute the inter-annotator agreement. We measured an average pairwise micro-averaged F1 score of 97.7\% for the inter-annotator agreement, showing the reliability of the annotation process.




\subsection{Stage 3: Relation annotation and declaration generation}
\begin{figure*}[h!]
\centering\includegraphics[width=2\columnwidth]{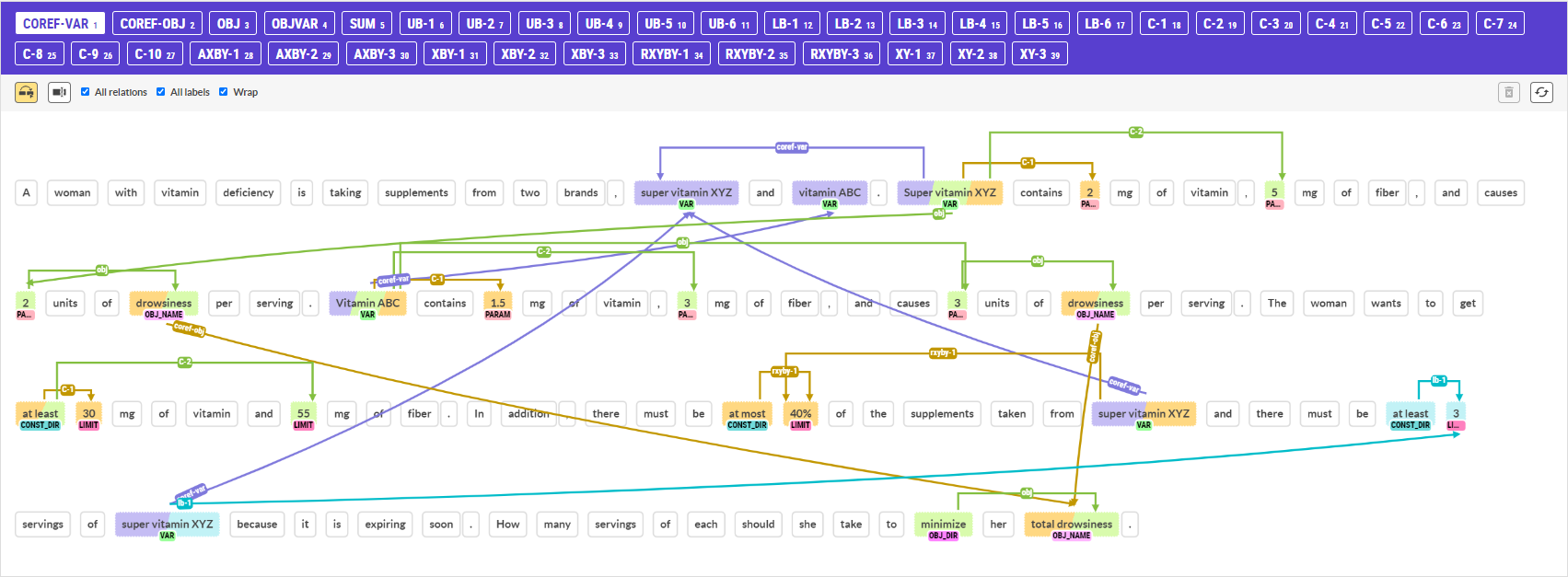}
\caption{Screenshot of REL annotation using Prodigy.}
\label{appendix:fig:rel-annotation-screenshot}
\end{figure*}

In the final stage, we annotated relations between entities for representing the objective and constraint declarations. Using a custom Prodigy annotation recipe, we integrated some validation checks to detect mistakes in the relation annotation period. These checks ensured that the annotator corrected all mistakes before proceeding to the next problem. Table \ref{rel-annotation-table} summarizes the relations required to represent each constraint type. The validation checks verified the number, labels, directions, and the entity labels of all relations that represent one constraint or objective.
To increase the efficiency of the annotation process, we also provided guidelines and trained annotators to annotate each type of constraints. Similar to stage 2, we worked with annotators to resolve complex relations. A screenshot of a relation annotation session is provided in Figure \ref{appendix:fig:rel-annotation-screenshot}.

After annotating the relations, we used a Python script to automatically convert the extracted relations into optimization declarations. A team of 5 people were asked to verify the correctness of the optimization declarations for all problems.  

\section{Examples of target domain problems}
Tables \ref{appendix:table:productiondomain}, \ref{appendix:table:transportationdomain} and \ref{appendix:table:healthsciencedomain} provide additional examples from the target domain: production, transportation, and sciences respectively.

\section{Predicted vs gold IR for target domain examples}
Table \ref{supp:ood-01}-\ref{supp:ood-7} demonstrate the predictions of our model in comparison to the gold standard (ground-truth) for several examples from the target domain. We show examples for when the model produced the gold IR formulation in Table \ref{supp:ood-01} and Table \ref{supp:ood-05}. More importantly, we illustrate and analyze different errors when the model was used to parse examples from the target domain. For each example, detailed analysis of the errors are given in the captions of Table \ref{supp:ood-02}-\ref{supp:ood-7}.

\section{Qualitative comparison of test predictions}
In Table \ref{appendix:table:furnituretable} and \ref{appendix:table:flowertable}, we qualitatively compare the generated IR of two different optimization problems. Our model perfectly matched the gold IR for both problems. When the copy mechanism is not used, the model made few errors when generating the constraint declarations. In the first example (Table \ref{appendix:table:furnituretable}), the model hallucinates the wrong constraint limits in the first two constraints. Furthermore, the constraint type is invalid for the second constraint. Similar errors happen in the problem of Table \ref{appendix:table:flowertable}.

\section{Experimental settings}
\paragraph{NER experiments.} We trained the XLM-RoBERTa transformer as the baseline model. The training was performed as a two-step approach with a training followed by a fine-tuning step. The training step utilized the HuggingFace \lstinline{get_linear_schedule_with_warmup} function that uses a learning rate decreasing linearly to 0 from the initial learning rate (1E-4) after a warmup period. This step was run for a maximum of 25 epochs on a batch of 64 samples with the early stopping callback function set to stop training monitoring the loss of the development split with a patience of 5 epochs and minimum change of 0.001. A model checkpoint was also used to save a checkpoint of the model that performed the best on the development set. The fine-tuning step also utilized a learning rate of 0.0001 for a maximum of 30 epochs with the same callback functions used in the training step. To set the learning rate, we used a grid search using a development set.

\paragraph{Generation experiments.} We trained the BART baseline model and our model for a total of 200 epochs, using a learning rate of 1E-06, and with a batch size of 32. The corresponding performance on the dev set for the best model is summarized in Table \ref{appendix:tab:dev}. We trained the Text-to-Table for a total of 4000 updates, using a learning rate of 1E-05, and with max-tokens of 4096. To set the learning rate, we used a grid search using a development set.

\newpage

\onecolumn
\begin{longtable}{|p{4.5cm}|p{10.5cm}|}
\hline
Problem Description & 
A hotel employs cleaners and receptionists. Cleaners earn \$500 per week and receptionists earn \$350 per week. The hotel requires a minimum of 100 workers of whom at least 20 must be receptionists. To keep the hotel clean and running smoothly, the number of receptionists should be at least a third of the number of cleaners. The hotel wants to keep the weekly wage bill below \$30000. Formulate a LP to minimize the wage bill.
\\
\hline
Text-to-Table + Prompt Form & 
\multicolumn{1}{|l|}{
\begin{tabular}{lrrr}
 & cleaners & receptionists & rhs \\
 objective & 500.0 & 350.0 &  \\
minimum & -1.0 & -1.0 & -100.0 \\
at least & 0.0 & -1.0 & -20.0\\
at least & 0.3333 & -1.0 & 0.0 \\
below & 500.0 & 350.0 & 30000.0 
\end{tabular}}
\\
\hline
Canonical Form & 
\multicolumn{1}{|l|}{
\begin{tabular}{lrrr}
 & var\_0 & var\_1 & rhs  \\
 objective & 500.0 & 350.0 &  \\
 constraint\_0 & -1.0 & -1.0 & -100.0 \\
 constraint\_1 & 0.0 & -1.0 & -20.0\\
 constraint\_2 & 0.3333 & -1.0 & 0.0 \\
 constraint\_3 & 500.0 & 350.0 & 30000.0 
\end{tabular}}
\\
\hline
\caption{Text-to-Table example: Problem description, and expected output. Newline tokens have been replaced with newline characters, indentation was added, and column dividers were omitted for readability.}
\label{appendix:table:texttotable}
\end{longtable}

\begin{table}[h!]
\centering\includegraphics[width=1\columnwidth]{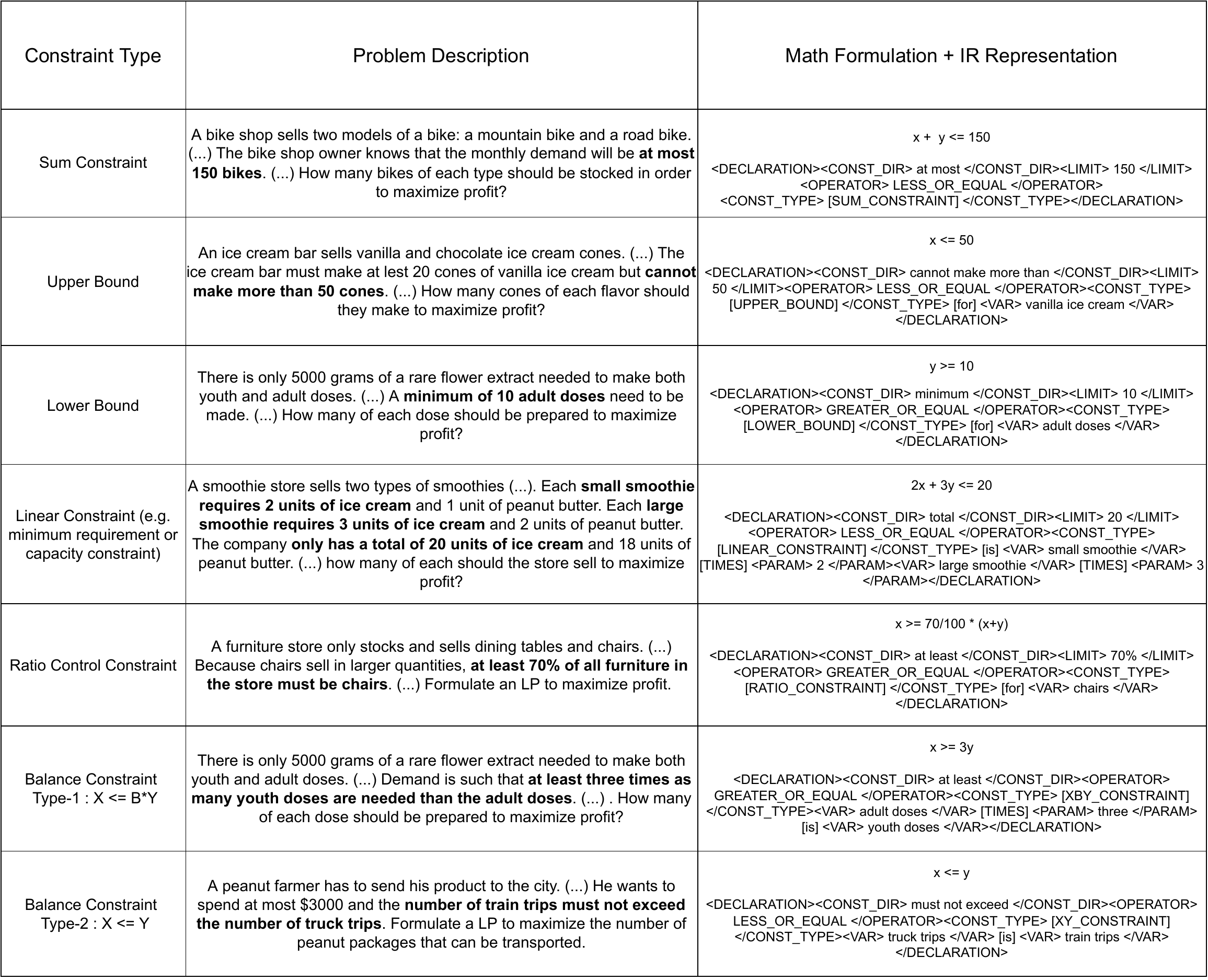}
\caption{Different types of constraints with examples from the LPWP dataset.}
\label{constraint-table}
\end{table}

\begin{table}[h!]
\centering\includegraphics[width=1\columnwidth]{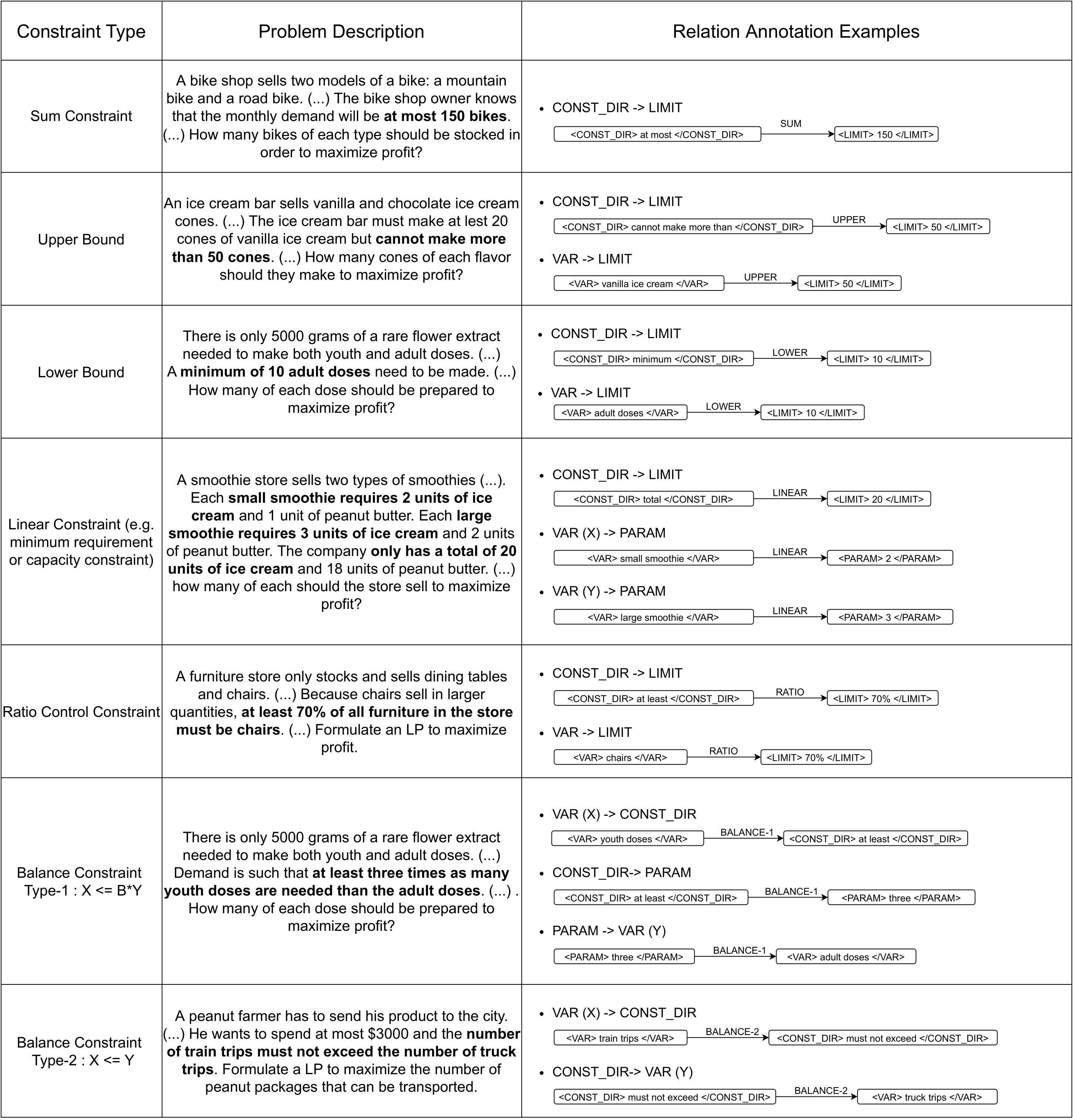}
\caption{REL annotation examples for different constraint types.}
\label{rel-annotation-table}
\end{table}

\onecolumn
\begin{longtable}{|p{4.5cm}|p{10.5cm}|}
\hline
Problem Description & 
There is only 5000 grams of a rare flower extract needed to make both youth and adult doses. Youth doses contain 20 grams of extract and adult doses contain 35 grams. Demand is such that at least three times as many youth doses are needed than the adult doses. A minimum of 10 adult doses need to be made. Youth doses are sold for a profit of \$5 while adult doses are sold at a profit of \$3. How many of each dose should be prepared to maximize profit?
\\
\hline
Intermediate Representation & 
\begin{lstlisting}
<DECLARATION>
	<OBJ_DIR> maximize </OBJ_DIR>
	<OBJ_NAME> profit </OBJ_NAME> [is] 
	<VAR> Youth doses </VAR> [TIMES] <PARAM> 5 </PARAM>
	<VAR> adult doses </VAR> [TIMES] <PARAM> 3 </PARAM>
</DECLARATION>

<DECLARATION>
	<CONST_DIR> only </CONST_DIR><LIMIT> 5000 </LIMIT>
	<OPERATOR> LESS_OR_EQUAL </OPERATOR> 
	<CONST_TYPE> [LINEAR_CONSTRAINT] </CONST_TYPE> [is]
	<VAR> Youth doses </VAR> [TIMES] <PARAM> 20 </PARAM>
	<VAR> adult doses </VAR> [TIMES] <PARAM> 35 </PARAM>
</DECLARATION>

DECLARATION>
	<CONST_DIR> at least </CONST_DIR>
	<OPERATOR> GREATER_OR_EQUAL </OPERATOR>
	<CONST_TYPE> [XBY_CONSTRAINT] </CONST_TYPE>
	<VAR> adult doses </VAR> [TIMES] 
	<PARAM> three </PARAM> [is] 
	<VAR> youth doses </VAR>
</DECLARATION>

<DECLARATION>
	<CONST_DIR> minimum </CONST_DIR><LIMIT> 10 </LIMIT>	
	<OPERATOR> GREATER_OR_EQUAL </OPERATOR>
	<CONST_TYPE> [LOWER_BOUND] </CONST_TYPE> [for]
	<VAR> adult doses </VAR>
</DECLARATION>
\end{lstlisting}

\\
\hline
Canonical Form & 
\multicolumn{1}{|l|}{
\begin{tabular}{lrrr}
 & var\_0 & var\_1 & rhs \\
 objective & 5 & 3 &  \\
 constraint\_0 & 20 & 35 & 5000 \\
 constraint\_1 & -1 & 3 & 0\\
 constraint\_2 & 0 & -1.0 & -10
\end{tabular}}
\\
\hline
Math Formulation &
\begin{lstlisting}
max 5x + 3y
subject to
    20x + 35y <= 5000
    x >= 3y
    y >= 10
\end{lstlisting}\\
\hline
\caption{Original dataset - Resource allocation example: problem description, intermediate representation, canonical form, and math formulation.}
\label{appendix:table:resourcedomain}
\end{longtable}

\onecolumn
\begin{longtable}{|p{4.5cm}|p{10.5cm}|}
\hline
Problem Description & 
Your client has \$60,000 available to invest for a 1 year term. The money can be placed in a trust yielding a 2\% return or in a savings account yielding a 3\% return. Based on your experience, you advise your client that at least 15\% of the investment be placed in the trust and that at most 80\% of the investment be placed in the savings account. How much should your client invest in each so as to maximize his return on investment?
\\
\hline
Intermediate Representation & 
\begin{lstlisting}
<DECLARATION>
    <OBJ_DIR> maximize </OBJ_DIR>
    <OBJ_NAME> return </OBJ_NAME> [is] 
    <VAR> trust </VAR> [TIMES] <PARAM> 2% </PARAM>
    <VAR> savings account </VAR> [TIMES] <PARAM> 3% </PARAM>
</DECLARATION>

<DECLARATION>
    <CONST_DIR> available </CONST_DIR><LIMIT> 60,000 </LIMIT>
    <OPERATOR> LESS_OR_EQUAL </OPERATOR>
    <CONST_TYPE> [SUM_CONSTRAINT] </CONST_TYPE>
</DECLARATION>

<DECLARATION>
    <CONST_DIR> at least </CONST_DIR><LIMIT> 15% </LIMIT>
    <OPERATOR> GREATER_OR_EQUAL </OPERATOR>
    <CONST_TYPE> [RATIO_CONSTRAINT] </CONST_TYPE> [for] 
    <VAR> trust </VAR>
</DECLARATION>

<DECLARATION>
    <CONST_DIR> at most </CONST_DIR><LIMIT> 80% </LIMIT>
    <OPERATOR> LESS_OR_EQUAL </OPERATOR>
    <CONST_TYPE> [RATIO_CONSTRAINT] </CONST_TYPE> [for] 
    <VAR> savings account </VAR>
</DECLARATION>
\end{lstlisting}
\\
\hline
Canonical Form & 
\multicolumn{1}{|l|}{
\begin{tabular}{lrrr}
 & var\_0 & var\_1 & rhs  \\
 objective & 0.02 & 0.03 &  \\
 constraint\_0 & 1 & 1 & 60000 \\
 constraint\_1 & -0.85 & 0.15 & 0\\
 constraint\_2 & -0.8 & 0.2 & 0
\end{tabular}}
\\
\hline
Math Formulation & 
\begin{lstlisting}
max (2/100)*x + (3/100)*y
subject to
    x + y <= 60000
    x >= (15/100)*(x+y)
    y <= (80/100)*(x+y)
\end{lstlisting}\\
\hline
\caption{Original dataset - Investment allocation example: problem description, intermediate representation, canonical form, and math formulation.}
\label{appendix:table:investmentdomain}
\end{longtable}

\onecolumn
\begin{longtable}{|p{4.5cm}|p{10.5cm}|}
\hline
Problem Description & 
A farmer has 500 acres of land to grow turnips and pumpkins. Turnips require 50 minutes of watering and \$80 worth of pesticide per acre. Pumpkins require 90 minutes of watering and \$50 worth of pesticide per acre. The farmer has 40000 minutes available for watering and \$34000 available to spend on pesticide. If the revenue per acre of turnips is \$300 and the revenue per acre of pumpkins is \$450, how many acres of each should he grow to maximize his revenue.
\\
\hline
Intermediate Representation & 
\begin{lstlisting}
<DECLARATION>
    <OBJ_DIR> maximize </OBJ_DIR>
    <OBJ_NAME> revenue </OBJ_NAME> [is] 
    <VAR> turnips </VAR> [TIMES] <PARAM> 300 </PARAM>
    <VAR> pumpkins </VAR> [TIMES] <PARAM> 450 </PARAM>
</DECLARATION>

<DECLARATION>
    <CONST_DIR> has </CONST_DIR><LIMIT> 500 </LIMIT>
    <OPERATOR> LESS_OR_EQUAL </OPERATOR>
    <CONST_TYPE> [SUM_CONSTRAINT] </CONST_TYPE>
</DECLARATION>

<DECLARATION>
    <CONST_DIR> available </CONST_DIR><LIMIT> 40000 </LIMIT>
    <OPERATOR> LESS_OR_EQUAL </OPERATOR> 
    <CONST_TYPE> [LINEAR_CONSTRAINT] </CONST_TYPE> [is] 
    <VAR> Turnips </VAR> [TIMES] <PARAM> 50 </PARAM>
    <VAR> Pumpkins </VAR> [TIMES] <PARAM> 90 </PARAM>
</DECLARATION>

<DECLARATION>
    <CONST_DIR> available </CONST_DIR><LIMIT> 34000 </LIMIT>
    <OPERATOR> LESS_OR_EQUAL </OPERATOR> 
    <CONST_TYPE> [LINEAR_CONSTRAINT] </CONST_TYPE> [is] 
    <VAR> Turnips </VAR> [TIMES] <PARAM> 80 </PARAM>
    <VAR> Pumpkins </VAR> [TIMES] <PARAM> 50 </PARAM>
</DECLARATION>
\end{lstlisting}
\\
\hline
Canonical Form & 
\multicolumn{1}{|l|}{
\begin{tabular}{lrrr}
 & var\_0 & var\_1 & rhs \\
 objective & 300 & 450 &  \\
 constraint\_0 & 1 & 1 & 500 \\
 constraint\_1 & 50 & 90 & 40000\\
 constraint\_2 & 80 & 50 & 34000
\end{tabular}}
\\
\hline
Math Formulation &
\begin{lstlisting}
max 300x + 450y
subject to
    x + y <= 500
    50x + 90y <= 40000
    80x + 50y <= 34000
\end{lstlisting}\\
\hline
\caption{Original Dataset - Farming example: problem description, intermediate representation, canonical form, and math formulation.}
\label{appendix:table:farmingdomain}
\end{longtable}

\onecolumn
\begin{longtable}{|p{4.5cm}|p{10.5cm}|}
\hline
Problem Description & 
A mining company has available a total of 100 square miles of mining sites and considering the use of two mining techniques: heap leaching and vat leaching. For each square mile of land, heap leaching technique can have a daily production of 3 tons of rare earth oxide per square miles but it also creates 8 tons of polluted wastewater and requires 10 extraction machines. On the other hand, vat leaching technique produces 5 tons of rare earth oxide per square miles per day while creating 17 tons of polluted wastewater and requiring 20 extraction machines. There are 100 machines available and due to environmental regulations, the amount of polluted wastewater must be at most 90 tons daily. Find the proportion of lands that use each mining technique in order to maximize the daily production of rare earth oxide.
\\
\hline
Intermediate Representation & 
\begin{lstlisting}
<DECLARATION>
    <OBJ_DIR> maximize </OBJ_DIR>
    <OBJ_NAME> rare earth oxide </OBJ_NAME> [is] 
    <VAR> heap leaching </VAR> [TIMES] <PARAM> 3 </PARAM>
    <VAR> vat leaching </VAR> [TIMES] <PARAM> 5 </PARAM>
</DECLARATION>

<DECLARATION>
    <CONST_DIR> available </CONST_DIR><LIMIT> 100 </LIMIT>
    <OPERATOR> LESS_OR_EQUAL </OPERATOR>
    <CONST_TYPE> [LINEAR_CONSTRAINT] </CONST_TYPE> [is] 
    <VAR> heap leaching </VAR> [TIMES] <PARAM> 10 </PARAM>
    <VAR> vat leaching </VAR> [TIMES] <PARAM> 20 </PARAM>
</DECLARATION>

<DECLARATION>
    <CONST_DIR> at most </CONST_DIR><LIMIT> 90 </LIMIT>
    <OPERATOR> LESS_OR_EQUAL </OPERATOR>
    <CONST_TYPE> [LINEAR_CONSTRAINT] </CONST_TYPE> [is] 
    <VAR> heap leaching </VAR> [TIMES] <PARAM> 8 </PARAM>
    <VAR> vat leaching </VAR> [TIMES] <PARAM> 17 </PARAM>
</DECLARATION>

<DECLARATION>
    <CONST_DIR> available </CONST_DIR><LIMIT> 100 </LIMIT>
    <OPERATOR> LESS_OR_EQUAL </OPERATOR>
    <CONST_TYPE> [SUM_CONSTRAINT] </CONST_TYPE>
</DECLARATION>
\end{lstlisting}
\\
\hline
Canonical Form & 
\multicolumn{1}{|l|}{
\begin{tabular}{lrrr}
 & var\_0 & var\_1 & rhs \\
 objective & 3 & 5 &  \\
 constraint\_0 & 10 & 20 & 100 \\
 constraint\_1 & 8 & 17 & 90\\
 constraint\_2 & 1 & 1 & 100
\end{tabular}}
\\
\hline
Math Formulation &
\begin{lstlisting}
max 3x + 5y
subject to
    10x + 20y <= 100
    8x + 17y <= 90
    x + y <= 100
\end{lstlisting}\\
\hline
\caption{Out-of-domain dataset - Production example: problem description, intermediate representation, canonical form, and math formulation.}
\label{appendix:table:productiondomain}
\end{longtable}

\onecolumn
\begin{longtable}{|p{4.5cm}|p{10.5cm}|}
\hline
Problem Description & 
A shipping company need to transport packages by either truck or car. A truck can transport 50 packages per trip while a car can transport 30 packages per trip. In addition, a truck uses 20 liters of gas per trip while a car uses 15 liters of gas per trip. There can be at most 5 truck trips made and at least 30\% of all the trips must be made by car. The company needs to transport at least 500 packages. How many of each transportation should they use to minimize the total amount of gas consumed?
\\
\hline
Intermediate Representation & 
\begin{lstlisting}
<DECLARATION>
    <OBJ_DIR> minimize </OBJ_DIR>
    <OBJ_NAME> amount of gas </OBJ_NAME> [is] 
    <VAR> truck </VAR> [TIMES] <PARAM> 20 </PARAM>
    <VAR> car </VAR> [TIMES] <PARAM> 15 </PARAM>
</DECLARATION>

<DECLARATION>
    <CONST_DIR> at most </CONST_DIR><LIMIT> 5 </LIMIT>
    <OPERATOR> LESS_OR_EQUAL </OPERATOR>
    <CONST_TYPE> [UPPER_BOUND] </CONST_TYPE> [for] 
    <VAR> truck </VAR>
</DECLARATION>

<DECLARATION>
    <CONST_DIR> at least </CONST_DIR><LIMIT> 30% </LIMIT>
    <OPERATOR> GREATER_OR_EQUAL </OPERATOR>
    <CONST_TYPE> [RATIO_CONSTRAINT] </CONST_TYPE> [for] 
    <VAR> car </VAR>
</DECLARATION>

<DECLARATION>
    <CONST_DIR> at least </CONST_DIR><LIMIT> 500 </LIMIT>
    <OPERATOR> GREATER_OR_EQUAL </OPERATOR>
    <CONST_TYPE> [LINEAR_CONSTRAINT] </CONST_TYPE> [is] 
    <VAR> truck </VAR> [TIMES] <PARAM> 50 </PARAM>
    <VAR> car </VAR> [TIMES] <PARAM> 30 </PARAM>
</DECLARATION>
\end{lstlisting}
\\
\hline
Canonical Form & 
\multicolumn{1}{|l|}{
\begin{tabular}{lrrr}
 & var\_0 & var\_1 & rhs \\
 objective & 20 & 15 &  \\
 constraint\_0 & 1 & 0 & 5 \\
 constraint\_1 & 0.3 & -0.7 & 0\\
 constraint\_2 & -50 & -30 & -500
\end{tabular}}
\\
\hline
Math Formulation &
\begin{lstlisting}
min 20x + 15y
subject to
    x <= 5
    y >= (30/100)*(x+y)
    50x + 30y >= 500
\end{lstlisting}\\
\hline
\caption{Out-of-domain dataset - Transportation example: problem description, intermediate representation, canonical form, and math formulation.}
\label{appendix:table:transportationdomain}
\end{longtable}

\onecolumn
\begin{longtable}{|p{4.5cm}|p{10.5cm}|}
\hline
Problem Description & 
A patient is undergoing radiation treatment involving two beams, Beam 1 and Beam 2. Beam 1 delivers a dose of 0.3 units of medicine per minute to the benign area of the pancreas and 0.2 units of medicine per minute to the benign area of the skin. Beam 2 delivers 0.2 units of medicine per minute to the benign area of the pancreas and 0.1 units of medicine per minute to the benign area of the skin.  In addition, beam 1 delivers 0.6 units of medicine per minute to the tumor and beam 2 delivers 0.4 units of medicine per minute to the tumor. At most 4 units of medicine should be received by the skin and at least 3 units of medicine should be delivered to the tumor.  How many minutes of each beam should be used to minimize the total radiation received by the pancreas? 
\\
\hline
Intermediate Representation & 
\begin{lstlisting}
<DECLARATION>
    <OBJ_DIR> minimize </OBJ_DIR>
    <OBJ_NAME> total radiation received by the pancreas </OBJ_NAME> [is] 
    <VAR> Beam 1 </VAR> [TIMES] <PARAM> 0.3 </PARAM>
    <VAR> Beam 2 </VAR> [TIMES] <PARAM> 0.2 </PARAM>
</DECLARATION>

<DECLARATION>
    <CONST_DIR> At most </CONST_DIR><LIMIT> 4 </LIMIT>
    <OPERATOR> LESS_OR_EQUAL </OPERATOR>
    <CONST_TYPE> [LINEAR_CONSTRAINT] </CONST_TYPE> [is] 
    <VAR> Beam 1 </VAR> [TIMES] <PARAM> 0.2 </PARAM>
    <VAR> Beam 2 </VAR> [TIMES] <PARAM> 0.1 </PARAM>
</DECLARATION>

<DECLARATION>
    <CONST_DIR> at least </CONST_DIR><LIMIT> 3 </LIMIT>
    <OPERATOR> GREATER_OR_EQUAL </OPERATOR> 
    <CONST_TYPE> [LINEAR_CONSTRAINT] </CONST_TYPE> [is] 
    <VAR> beam 1 </VAR> [TIMES] <PARAM> 0.6 </PARAM>
    <VAR> beam 2 </VAR> [TIMES] <PARAM> 0.4 </PARAM>
</DECLARATION>
\end{lstlisting}
\\
\hline
Canonical Form & 
\multicolumn{1}{|l|}{
\begin{tabular}{lrrr}
 & var\_0 & var\_1 & rhs \\
 objective & 0.3 & 0.2 &  \\
 constraint\_0 & 0.2 & 0.1 & 4 \\
 constraint\_1 & -0.6 & -0.4 & 3
\end{tabular}}
\\
\hline
Math Formulation &
\begin{lstlisting}
min 0.3x + 0.2y
subject to
    0.2x + 0.1y <= 4
    0.6x + 0.4y >= 3
\end{lstlisting}\\
\hline
\caption{Out-of-domain dataset - Health Science example: problem description, intermediate representation, canonical form, and math formulation.}
\label{appendix:table:healthsciencedomain}
\end{longtable}

\onecolumn
\begin{table}[h!]
\begin{tabular}{|p{7.5cm}|p{7.5cm}|}
 \hline
 \multicolumn{2}{|p{15cm}|}{A furniture store only stocks and sells dining tables and chairs. The profit per dining table is \$350 and the profit per chair is \$75. There is 500 sq ft of space available and a dining table requires 8 sq ft of floor space while a chair requires 2 sq ft. Because chairs sell in larger quantities, at least 70\% of all furniture in the store must be chairs. In terms of capital, a dining table ties up \$1000 in capital and a chair ties up \$150 in capital. The company wants a maximum of \$20000 worth of capital tied up at any time. Formulate an LP to maximize profit.} \\
 \hline
  \multicolumn{1}{|c|}{OptGen w/o copy}&  \multicolumn{1}{|c|}{OptGen} \\
 \hline
 \begin{lstlisting}
 <DECLARATION><OBJ_DIR> maximize </OBJ_DIR><OBJ_NAME> profit </OBJ_NAME> [is] <VAR> dining table </VAR> [TIMES] <PARAM> 350 </PARAM><VAR> chair </VAR> [TIMES] <PARAM> 75 </PARAM></DECLARATION>
 \end{lstlisting}
 &
 \begin{lstlisting}
<DECLARATION><OBJ_DIR> maximize </OBJ_DIR><OBJ_NAME profit </OBJ_NAME> [is] <VAR> dining table </VAR> [TIMES] <PARAM> 350 </PARAM><VAR> chair </VAR> [TIMES] <PARAM> 75 </PARAM></DECLARATION>
 \end{lstlisting}
 \\
 \vspace{-2em}
 \begin{lstlisting} 
<DECLARATION><CONST_DIR> available </CONST_DIR><LIMIT>(*@ \hlred{20000} @*)</LIMIT><OPERATOR> LESS_OR_EQUAL </OPERATOR><CONST_TYPE> [LINEAR_CONSTRAINT] </CONST_TYPE> [is] <VAR> dining table </VAR> [TIMES] <PARAM> 8 </PARAM><VAR> chair </VAR> [TIMES] <PARAM> 2 </PARAM></DECLARATION>
 \end{lstlisting}
 &
 \vspace{-2em}
 \begin{lstlisting}
<DECLARATION><CONST_DIR> available </CONST_DIR><LIMIT>(*@ \hl{500} @*)</LIMIT><OPERATOR> LESS_OR_EQUAL </OPERATOR><CONST_TYPE> [LINEAR_CONSTRAINT] </CONST_TYPE> [is] <VAR> dining table </VAR> [TIMES] <PARAM> 8 </PARAM><VAR> chair </VAR> [TIMES] <PARAM> 2 </PARAM></DECLARATION>
 \end{lstlisting}
 \\
  \vspace{-2em}
  \begin{lstlisting}
<DECLARATION><CONST_DIR> at least </CONST_DIR><LIMIT>(*@ \hlred{200000} @*)</LIMIT><OPERATOR> GREATER_OR_EQUAL </OPERATOR><CONST_TYPE>(*@ \hlred{[LINEAR\_CONSTRAINT]} @*)</CONST_TYPE>(*@ \hlred{[is] <VAR> dining table </VAR> [TIMES] <PARAM> 1000 </PARAM><VAR> chair </VAR> [TIMES] <PARAM> 150 </PARAM>} @*)</DECLARATION>
 \end{lstlisting}
 &
  \vspace{-2em}
 \begin{lstlisting}
<DECLARATION><CONST_DIR> at least </CONST_DIR><LIMIT>(*@ \hl{70\%} @*)</LIMIT><OPERATOR> GREATER_OR_EQUAL </OPERATOR><CONST_TYPE>(*@ \hl{[RATIO\_CONSTRAINT]} @*)</CONST_TYPE> (*@\hl{[for] <VAR> chairs </VAR>} @*)</DECLARATION>
 \end{lstlisting}
 \\
  \vspace{-2em}
   \begin{lstlisting}
<DECLARATION><CONST_DIR> maximum </CONST_DIR><LIMIT> 20000 </LIMIT><OPERATOR> LESS_OR_EQUAL </OPERATOR><CONST_TYPE> [LINEAR_CONSTRAINT] </CONST_TYPE> [is] <VAR> dining table </VAR> [TIMES] <PARAM> 1000 </PARAM><VAR> chair </VAR> [TIMES] <PARAM> 150 </PARAM></DECLARATION>
 \end{lstlisting}
 &
  \vspace{-2em}
 \begin{lstlisting}
<DECLARATION><CONST_DIR> maximum </CONST_DIR><LIMIT> 20000 </LIMIT> LESS_OR_EQUAL </OPERATOR><CONST_TYPE> [LINEAR_CONSTRAINT] </CONST_TYPE> [is] <VAR> dining table </VAR> [TIMES] <PARAM> 1000 </PARAM><VAR> chair </VAR> [TIMES] <PARAM> 150 </PARAM></DECLARATION>
 \end{lstlisting}
 \\
 \hline
\end{tabular}
\caption{Qualitative comparison of model predictions with and without copy mechanism. In this example, the model without copy mechanism produced erroneous IR declarations (highlighted in red) whereas our model produces perfect matches with the gold declarations.}
\label{appendix:table:furnituretable}
\end{table}

\onecolumn
\begin{table}[h!]
\begin{tabular}{|p{7.5cm}|p{7.5cm}|}
 \hline
 \multicolumn{2}{|p{15cm}|}{There is only 5000 grams of a rare flower extract needed to make both youth and adult doses. Youth doses contain 20 grams of extract and adult doses contain 35 grams. Demand is such that at least three times as many youth doses are needed than the adult doses. A minimum of 10 adult doses need to be made. Youth doses are sold for a profit of \$5 while adult doses are sold at a profit of \$3. How many of each dose should be prepared to maximize profit?} \\
 \hline
  \multicolumn{1}{|c|}{OptGen w/o copy}&  \multicolumn{1}{|c|}{OptGen} \\
 \hline
 \begin{lstlisting}
<DECLARATION><OBJ_DIR> maximize </OBJ_DIR><OBJ_NAME profit </OBJ_NAME> [is] <VAR> Youth doses </VAR> [TIMES] <PARAM> 5 </PARAM><VAR> adult doses </VAR> [TIMES] <PARAM> 3 </PARAM></DECLARATION>
 \end{lstlisting}
 &
 \begin{lstlisting}
<DECLARATION><OBJ_DIR> maximize </OBJ_DIR><OBJ_NAME profit </OBJ_NAME> [is] <VAR> Youth doses </VAR> [TIMES] <PARAM> 5 </PARAM><VAR> adult doses </VAR> [TIMES] <PARAM> 3 </PARAM></DECLARATION>
 \end{lstlisting}
 \\
 \vspace{-2em}
 \begin{lstlisting}
<DECLARATION><CONST_DIR> only </CONST_DIR><LIMIT>(*@ \hlred{200000} @*)</LIMIT><OPERATOR> LESS_OR_EQUAL </OPERATOR><CONST_TYPE> [LINEAR_CONSTRAINT] </CONST_TYPE> [is] <VAR> Youth doses </VAR> [TIMES] <PARAM> 20 </PARAM><VAR> adult doses </VAR> [TIMES] <PARAM> 35 </PARAM></DECLARATION>
 \end{lstlisting}
 &
 \vspace{-2em}
 \begin{lstlisting}
<DECLARATION><CONST_DIR> only </CONST_DIR><LIMIT>(*@ \hl{5000} @*)</LIMIT><OPERATOR> LESS_OR_EQUAL </OPERATOR><CONST_TYPE> [LINEAR_CONSTRAINT] </CONST_TYPE> [is] <VAR> Youth doses </VAR> [TIMES] <PARAM> 20 </PARAM><VAR> adult doses </VAR> [TIMES] <PARAM> 35 </PARAM></DECLARATION>
 \end{lstlisting}
 \\
  \vspace{-2em}
  \begin{lstlisting}
<DECLARATION><CONST_DIR> at least </CONST_DIR><OPERATOR> GREATER_OR_EQUAL </OPERATOR><CONST_TYPE> [XBY_CONSTRAINT] </CONST_TYPE><VAR>(*@ \hlred{youth doses} @*)</VAR> [TIMES] <PARAM> three </PARAM> [is] <VAR>(*@ \hlred{adult doses} @*)</VAR></DECLARATION>
 \end{lstlisting}
 &
  \vspace{-2em}
 \begin{lstlisting}
<DECLARATION><CONST_DIR> at least </CONST_DIR><OPERATOR> GREATER_OR_EQUAL </OPERATOR><CONST_TYPE> [XBY_CONSTRAINT] </CONST_TYPE><VAR>(*@ \hl{adult doses} @*)</VAR> [TIMES] <PARAM> three </PARAM> [is] <VAR>(*@ \hl{youth doses} @*)</VAR></DECLARATION>
 \end{lstlisting}
 \\
  \vspace{-2em}
   \begin{lstlisting}
<DECLARATION><CONST_DIR> minimum </CONST_DIR><LIMIT>(*@ \hlred{200000} @*)</LIMIT><OPERATOR> GREATER_OR_EQUAL </OPERATOR><CONST_TYPE>(*@ \hlred{[LINEAR\_CONSTRAINT]} @*)</CONST_TYPE>(*@ \hlred{[is] <VAR> Youth doses </VAR> [TIMES] <PARAM> 20 </PARAM><VAR> adult doses </VAR> [TIMES] <PARAM> 35 </PARAM>}@*)</DECLARATION>
 \end{lstlisting}
 &
  \vspace{-2em}
 \begin{lstlisting}
<DECLARATION><CONST_DIR> minimum </CONST_DIR><LIMIT>(*@ \hl{10} @*)</LIMIT><OPERATOR> GREATER_OR_EQUAL </OPERATOR><CONST_TYPE>(*@ \hl{[LOWER\_BOUND]} @*)</CONST_TYPE>(*@ \hl{[for] <VAR> adult doses </VAR>}@*)</DECLARATION>
 \end{lstlisting}
 \\
 \hline
\end{tabular}
\caption{Qualitative comparison of model predictions with and without copy mechanism. In this example, the model without copy mechanism produced erroneous IR declarations (highlighted in red) whereas our model produces perfect matches with the gold declarations.}
\label{appendix:table:flowertable}
\end{table}

\begin{figure}[h!]
\centering
\fontsize{10}{12}\selectfont
\centering
\subfloat[\centering Problem level errors on the Source domain]{{\includegraphics[width=\linewidth]{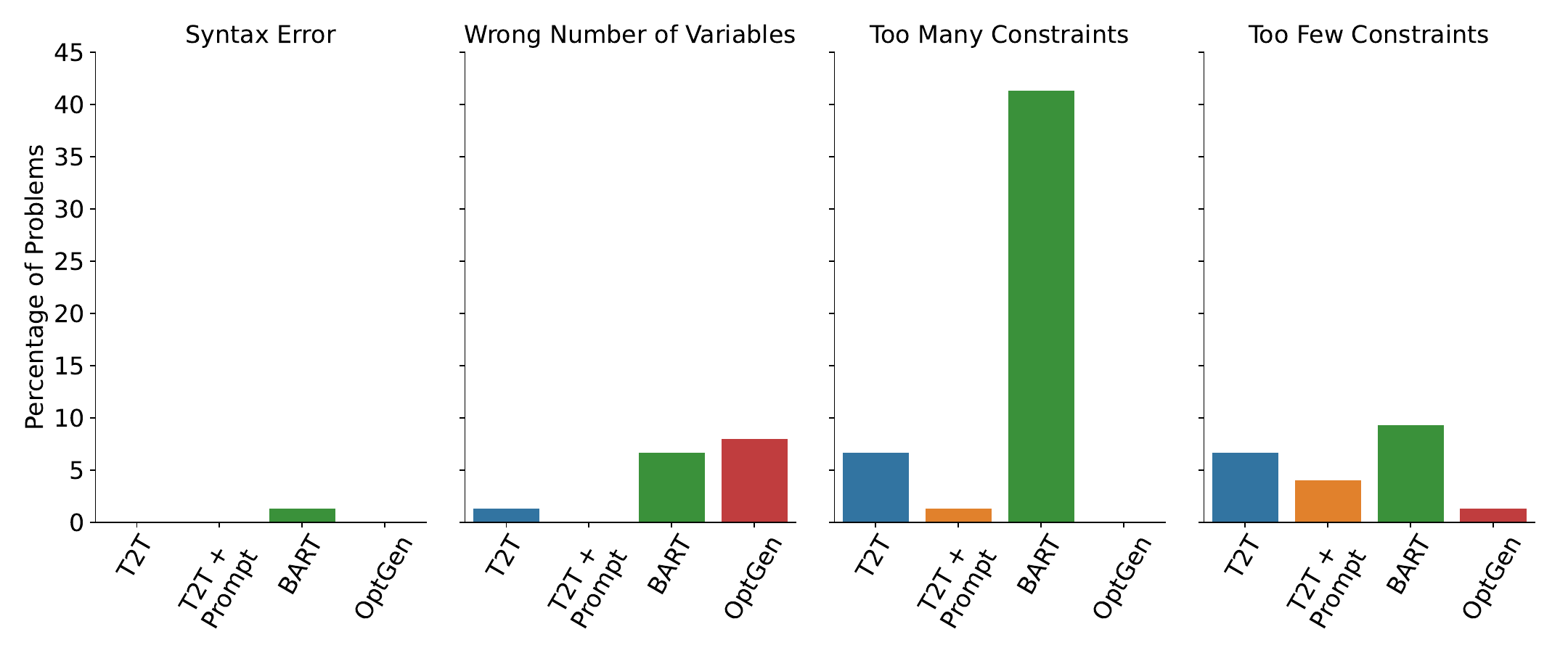} }}%
\qquad
\subfloat[\centering Problem level errors on the Target domain]{{\includegraphics[width=\linewidth]{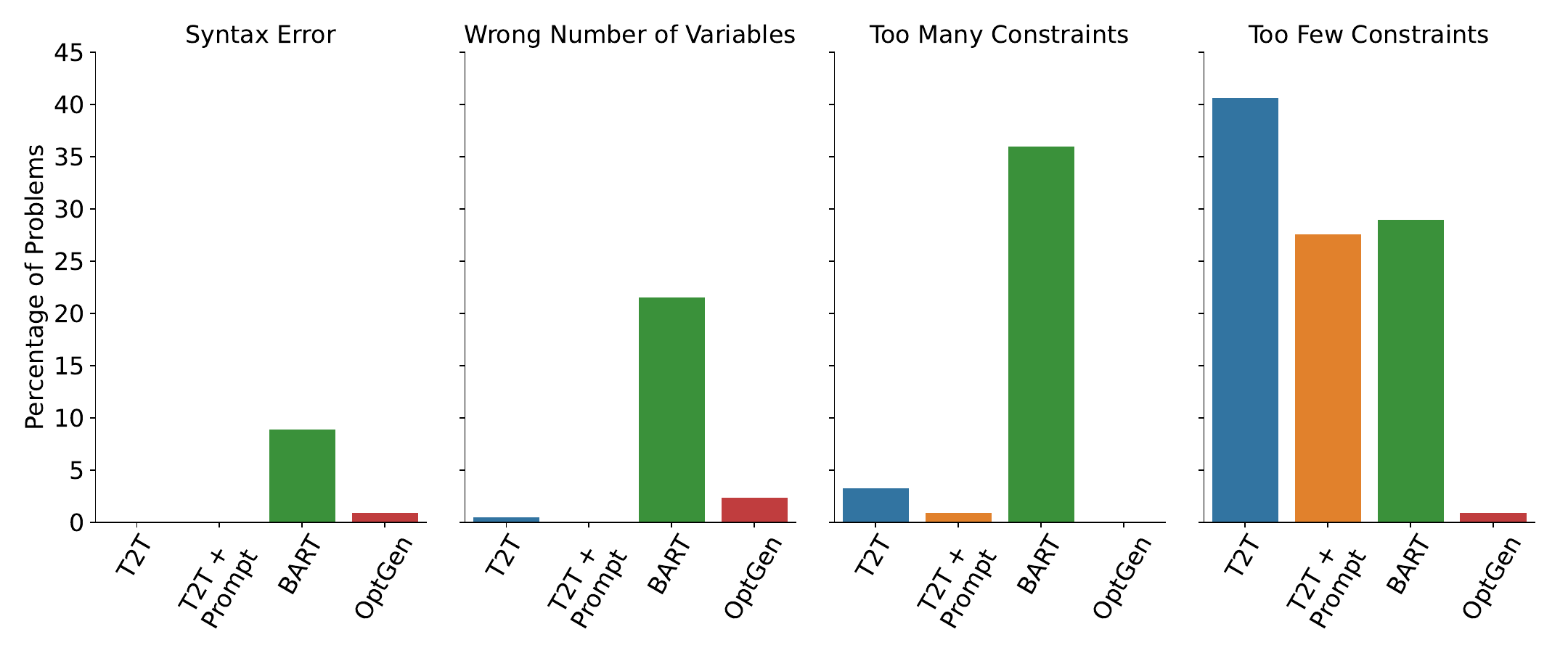} }}%
\caption{Classification of problem level errors for each model. These errors are not mutually exclusive. Problem level syntax errors will result in a completely incorrect problem as they cannot be parsed. On models predicting the IR of a problem, a problem level syntax error is defined as one that cannot be parsed through an XML parser. On the T2T models, syntax errors are defined as a mismatch between the number of columns in each row.}
\label{appendix:fig:problemlevelerrors}
\end{figure}

\begin{figure}[h!]
\centering
\fontsize{10}{12}\selectfont
\centering
\subfloat[\centering Declaration level errors on the Source domain.]{{\includegraphics[width=0.75\linewidth]{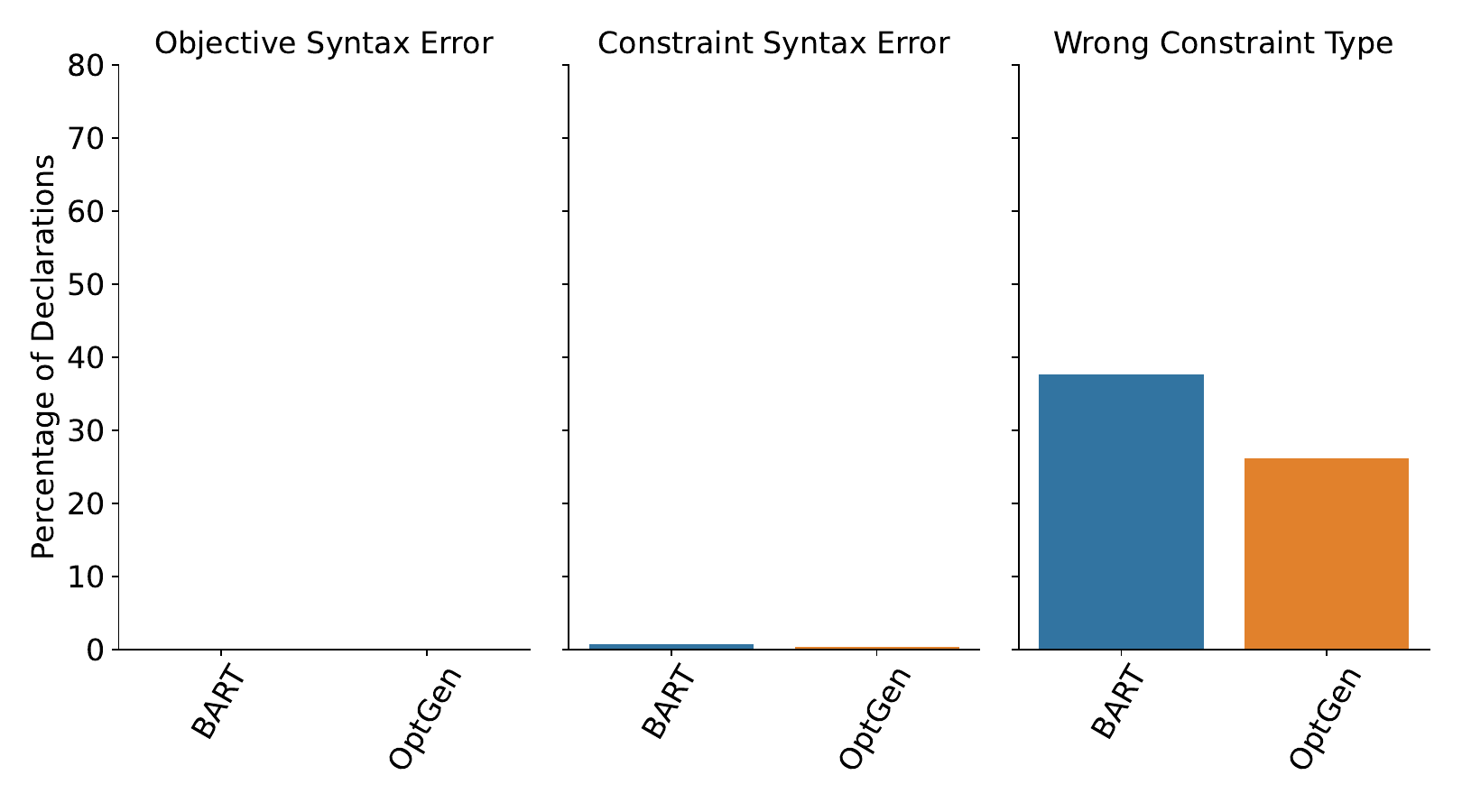}\includegraphics[width=0.25\linewidth]{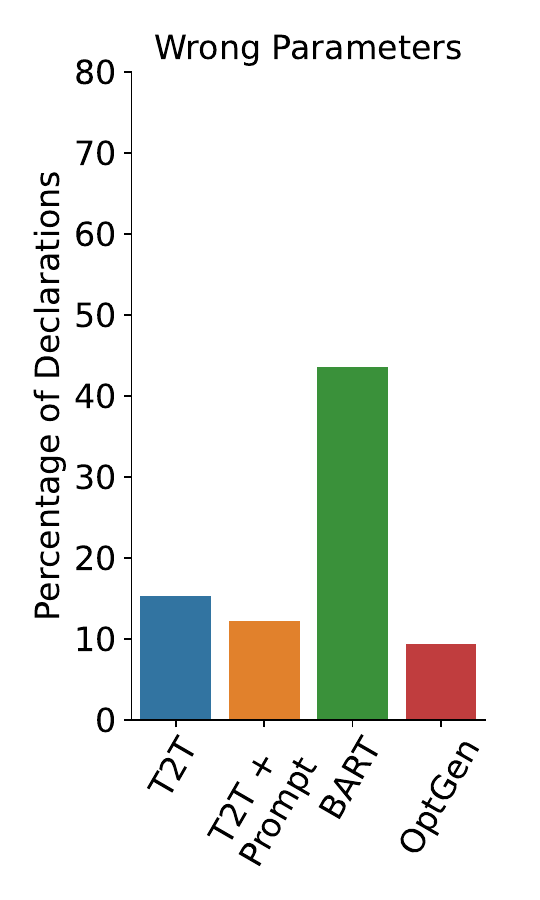}  }}%
\qquad
\subfloat[\centering Declaration level errors on the Target domain.]{{\includegraphics[width=0.75\linewidth]{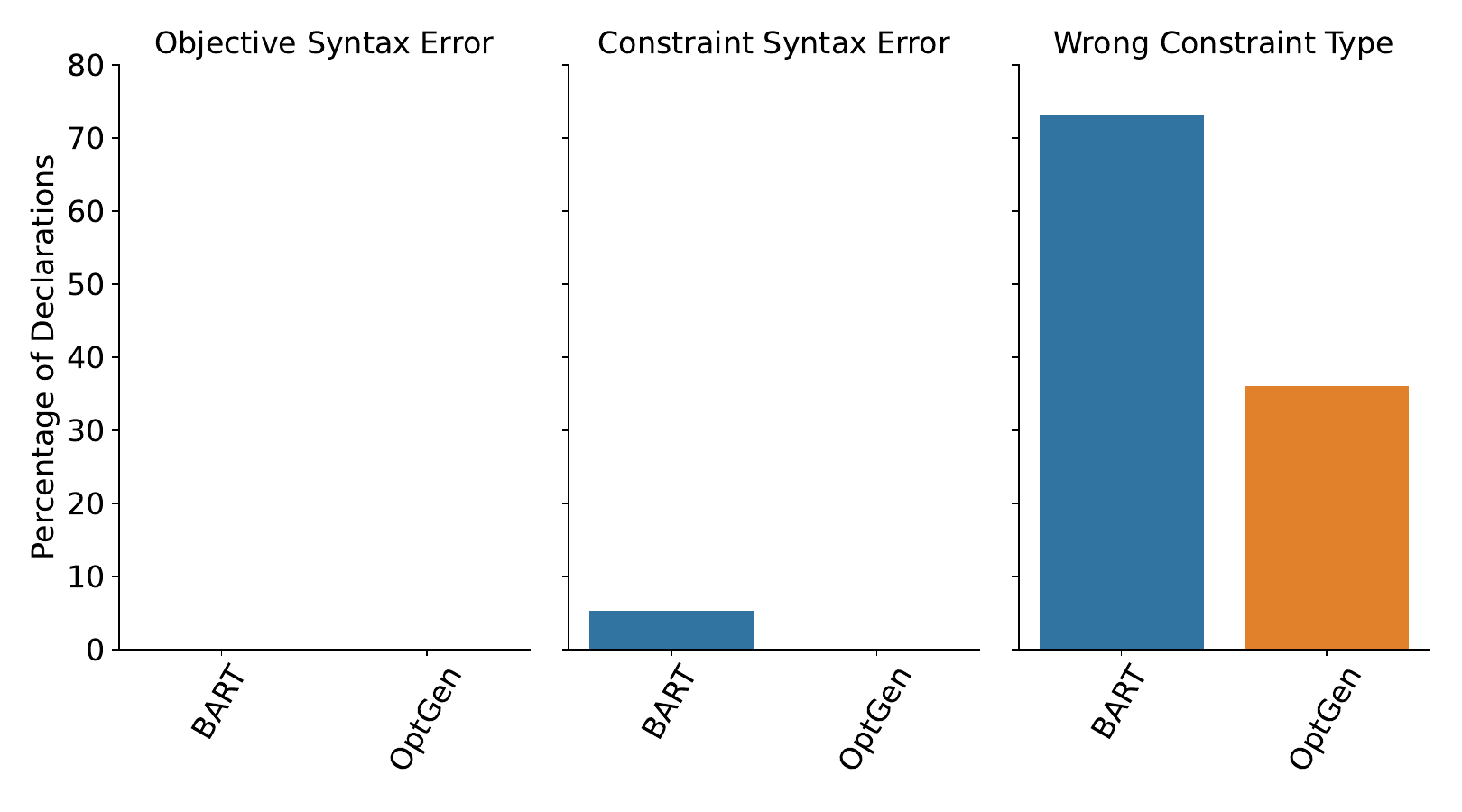}\includegraphics[width=0.25\linewidth]{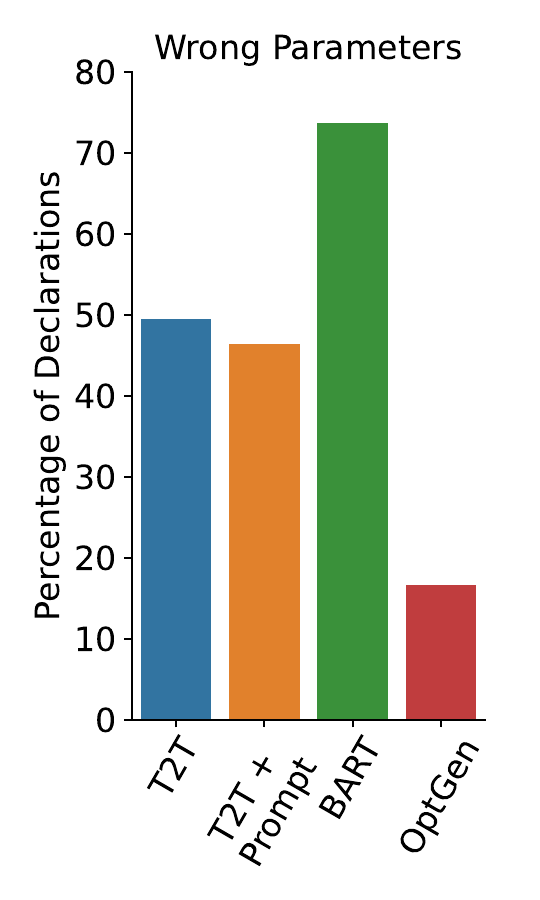} }}%
\caption{Classification of declaration level errors for each model. Note that some of these error types are not made by the T2T models, as they do not predict constraint types, and their syntax errors are all at the problem level. The parser will skip parsing poorly formatted declarations, which are later counted as incorrect.  We define a declaration level parameter error as a declaration that contains any parameter mismatch between the prediction and the ground truth.}
\label{appendix:fig:declarationlevelerrors}
\end{figure}


\onecolumn
\begin{table}[h!]
\begin{tabular}{|p{7.5cm}|p{7.5cm}|}
 \hline
 \multicolumn{2}{|p{15cm}|}{A patient in the hospital can take two pills, Pill 1 and Pill 2. Per pill, pill 1 provides 0.2 units of pain medication and 0.3 units of anxiety medication. Per pill, pill 2 provides 0.6 units of pain medication and 0.2 units of anxiety medication. In addition, pill 1 causes 0.3 units of discharge while pill 2 causes 0.1 units of discharge. At most 6 units of pain medication can be provided and at least 3 units of anxiety medication must be provided. How many pills of each should the patient be given to minimize the total amount of discharge?} \\
 \hline
  \multicolumn{1}{|c|}{Gold}&  \multicolumn{1}{|c|}{Pred} \\
 \hline
 \begin{lstlisting}
<DECLARATION><OBJ_DIR> minimize </OBJ_DIR><OBJ_NAME> amount of discharge </OBJ_NAME> [is] <VAR> pill 1 </VAR> [TIMES] <PARAM> 0.3 </PARAM><VAR> pill 2 </VAR> [TIMES] <PARAM> 0.1 </PARAM></DECLARATION>
 \end{lstlisting}
 &
 \begin{lstlisting}
<DECLARATION><OBJ_DIR> minimize </OBJ_DIR><OBJ_NAME>(*@ \hlred{total} @*)amount of discharge </OBJ_NAME> [is] <VAR> Pill 1 </VAR> [TIMES] <PARAM> 0.3 </PARAM><VAR> pill 2 </VAR> [TIMES] <PARAM> 0.1 </PARAM></DECLARATION>
 \end{lstlisting}
 \\
 \vspace{-2em}
 \begin{lstlisting}
<DECLARATION><CONST_DIR> At most </CONST_DIR><LIMIT> 6 </LIMIT><OPERATOR> LESS_OR_EQUAL </OPERATOR><CONST_TYPE> [LINEAR_CONSTRAINT] </CONST_TYPE> [is] <VAR> pill 1 </VAR> [TIMES] <PARAM> 0.2 </PARAM><VAR> pill 2 </VAR> [TIMES] <PARAM> 0.6 </PARAM></DECLARATION>
 \end{lstlisting}
 &
 \vspace{-2em}
 \begin{lstlisting}
<DECLARATION><CONST_DIR> At most </CONST_DIR><LIMIT> 6 </LIMIT><OPERATOR> LESS_OR_EQUAL </OPERATOR><CONST_TYPE> [LINEAR_CONSTRAINT] </CONST_TYPE> [is] <VAR> pill 1 </VAR> [TIMES] <PARAM> 0.2 </PARAM><VAR> pill 2 </VAR> [TIMES] <PARAM> 0.6 </PARAM></DECLARATION>
 \end{lstlisting}
 \\
  \vspace{-2em}
  \begin{lstlisting}
<DECLARATION><CONST_DIR> at least </CONST_DIR><LIMIT> 3 </LIMIT><OPERATOR> GREATER_OR_EQUAL </OPERATOR><CONST_TYPE> [LINEAR_CONSTRAINT] </CONST_TYPE> [is] <VAR> pill 1 </VAR> [TIMES] <PARAM> 0.3 </PARAM><VAR> pill 2 </VAR> [TIMES] <PARAM> 0.2 </PARAM></DECLARATION>
 \end{lstlisting}
 &
  \vspace{-2em}
 \begin{lstlisting}
<DECLARATION><CONST_DIR> at least </CONST_DIR><LIMIT> 3 </LIMIT><OPERATOR> GREATER_OR_EQUAL </OPERATOR><CONST_TYPE> [LINEAR_CONSTRAINT] </CONST_TYPE> [is] <VAR> pill 1 </VAR> [TIMES] <PARAM> 0.3 </PARAM><VAR> pill 2 </VAR> [TIMES] <PARAM> 0.2 </PARAM></DECLARATION>
 \end{lstlisting}
 \\
 \hline
\end{tabular}
\caption{Comparison of predicted vs. gold IR for out-of-domain example (Health Science). In this example, the predicted IR is almost equal to the gold IR except for the extra token ("total") in the objective name declaration.} \label{supp:ood-01}
\end{table}

\onecolumn
\begin{table}[h!]
\begin{tabular}{|p{7.5cm}|p{7.5cm}|}
 \hline
 \multicolumn{2}{|p{15cm}|}{An international goods exporter uses ships and planes to transport goods. A ship can take 40 containers worth of goods and uses 500 liters of fuel per trip. A plane can take 20 containers worth of goods and uses 300 liters of fuel per trip. The company needs to transport at least 500 containers worth of goods. In addition, there can be at most 10 plane trips made and a minimum of 50\% of the trips made must be by ship. How many of each trip should be made to minimize the total amount of fuel consumed?} \\
 \hline
  \multicolumn{1}{|c|}{Gold}&  \multicolumn{1}{|c|}{Pred} \\
 \hline
 \begin{lstlisting}
<DECLARATION><OBJ_DIR> minimize </OBJ_DIR><OBJ_NAME> total amount of fuel </OBJ_NAME> [is] <VAR> ship </VAR> [TIMES] <PARAM> 500 </PARAM><VAR> plane </VAR> [TIMES] <PARAM> 300 </PARAM></DECLARATION>
 \end{lstlisting}
 &
 \begin{lstlisting}
<DECLARATION><OBJ_DIR> minimize </OBJ_DIR><OBJ_NAME> total amount of fuel </OBJ_NAME> [is] <VAR> ship </VAR> [TIMES] <PARAM> 500 </PARAM><VAR> plane </VAR> [TIMES] <PARAM> 300 </PARAM></DECLARATION>
 \end{lstlisting}
 \\
 \vspace{-2em}
 \begin{lstlisting}
<DECLARATION><CONST_DIR> at least </CONST_DIR><LIMIT> 500 </LIMIT><OPERATOR> GREATER_OR_EQUAL </OPERATOR><CONST_TYPE> [LINEAR_CONSTRAINT] </CONST_TYPE> [is] <VAR> ship </VAR> [TIMES] <PARAM> 40 </PARAM><VAR> plane </VAR> [TIMES] <PARAM> 20 </PARAM></DECLARATION>
 \end{lstlisting}
 &
 \vspace{-2em}
 \begin{lstlisting}
<DECLARATION><CONST_DIR> at least </CONST_DIR><LIMIT> 500 </LIMIT><OPERATOR> GREATER_OR_EQUAL </OPERATOR><CONST_TYPE> [LINEAR_CONSTRAINT] </CONST_TYPE> [is] <VAR> ship </VAR> [TIMES] <PARAM> 40 </PARAM><VAR> plane </VAR> [TIMES] <PARAM> 20 </PARAM></DECLARATION>
 \end{lstlisting}
 \\
  \vspace{-2em}
  \begin{lstlisting}
<DECLARATION><CONST_DIR> at most </CONST_DIR><OPERATOR> LESS_OR_EQUAL </OPERATOR><LIMIT> 10 </LIMIT><CONST_TYPE> [UPPER_BOUND] </CONST_TYPE> [for] <VAR> plane </VAR></DECLARATION>
 \end{lstlisting}
 &
  \vspace{-2em}
 \begin{lstlisting}
<DECLARATION><CONST_DIR> at most </CONST_DIR><OPERATOR> LESS_OR_EQUAL </OPERATOR><LIMIT> 10 </LIMIT><CONST_TYPE> [UPPER_BOUND] </CONST_TYPE> [for] <VAR> plane </VAR></DECLARATION>
 \end{lstlisting}
 \\
  \vspace{-2em}
  \begin{lstlisting}
<DECLARATION><CONST_DIR> minimum </CONST_DIR><OPERATOR> GREATER_OR_EQUAL </OPERATOR><LIMIT> 50% </LIMIT><CONST_TYPE> [RATIO_CONSTRAINT] </CONST_TYPE> [for] <VAR> ship </VAR></DECLARATION>
 \end{lstlisting}
 &
  \vspace{-2em}
 \begin{lstlisting}
<DECLARATION><CONST_DIR> minimum </CONST_DIR><OPERATOR> GREATER_OR_EQUAL </OPERATOR><LIMIT> 50% </LIMIT><CONST_TYPE> [RATIO_CONSTRAINT] </CONST_TYPE> [for] <VAR> ship </VAR></DECLARATION>
 \end{lstlisting}
 \\
 \hline
\end{tabular}
\caption{Comparison of predicted vs. gold IR for out-of-domain example (Transportation example). In this example, the model perfectly matched the gold IR.}
\label{supp:ood-05}
\end{table}

\onecolumn
\begin{table}[ht]
\begin{tabular}{|p{7.5cm}|p{7.5cm}|}
 \hline
 \multicolumn{2}{|p{15cm}|}{A pharmacy has 3000 mg of morphine to make painkillers and sleeping pills. Each painkiller pill requires 10 mg of morphine and 3 units of digestive medicine. Each sleeping pill requires 6 mg of morphine and 5 units of digestive medicine. The pharmacy needs to make at least 50 painkiller pills. Since sleeping pills are more popular, at least 70\% of the pills should be sleeping pills. How many of each should the pharmacy make to minimize the total amount of digestive medicine needed?} \\
 \hline
  \multicolumn{1}{|c|}{Gold}&  \multicolumn{1}{|c|}{Pred} \\
 \hline
 \begin{lstlisting}
<DECLARATION><OBJ_DIR> minimize </OBJ_DIR><OBJ_NAME> amount of digestive medicine </OBJ_NAME> [is] <VAR> painkiller pill </VAR> [TIMES] <PARAM> 3 </PARAM><VAR> sleeping pill </VAR> [TIMES] <PARAM> 5 </PARAM></DECLARATION>
 \end{lstlisting}
 &
 \begin{lstlisting}
<DECLARATION><OBJ_DIR> minimize </OBJ_DIR><OBJ_NAME> amount of digestive medicine </OBJ_NAME> [is] <VAR> painkiller pill </VAR> [TIMES] <PARAM> 3 </PARAM><VAR> sleeping pill </VAR> [TIMES] <PARAM> 5 </PARAM></DECLARATION>
 \end{lstlisting}
 \\
 \vspace{-2em}
 \begin{lstlisting}
<DECLARATION><CONST_DIR> at least </CONST_DIR><LIMIT> 50 </LIMIT><OPERATOR> GREATER_OR_EQUAL </OPERATOR><CONST_TYPE> [LOWER_BOUND] </CONST_TYPE> [for] <VAR> painkiller pills </VAR></DECLARATION>
 \end{lstlisting}
 &
 \vspace{-2em}
 \begin{lstlisting}
<DECLARATION><CONST_DIR> at least </CONST_DIR><LIMIT> 50 </LIMIT><OPERATOR> GREATER_OR_EQUAL </OPERATOR><CONST_TYPE> [LOWER_BOUND] </CONST_TYPE> [for] <VAR> painkiller pills </VAR></DECLARATION>
 \end{lstlisting}
 \\
  \vspace{-2em}
  \begin{lstlisting}
<DECLARATION><CONST_DIR> at least </CONST_DIR><LIMIT>(*@ \hlblue{70\%} @*)</LIMIT><OPERATOR> GREATER_OR_EQUAL </OPERATOR><CONST_TYPE>(*@ \hlblue{[RATIO\_CONSTRAINT]} @*) </CONST_TYPE> [for] <VAR> (*@ \hlblue{sleeping pills} @*)</VAR></DECLARATION>
 \end{lstlisting}
 &
  \vspace{-2em}
 \begin{lstlisting}
<DECLARATION><CONST_DIR> at least </CONST_DIR><LIMIT>(*@ \hlred{50} @*)</LIMIT><OPERATOR> GREATER_OR_EQUAL </OPERATOR><CONST_TYPE>(*@  \hlred{[LOWER\_BOUND]} @*)</CONST_TYPE> [for] <VAR>(*@ \hlred{painkiller pills} @*)</VAR></DECLARATION>
 \end{lstlisting}
 \\
  \vspace{-2em}
  \begin{lstlisting}
<DECLARATION><CONST_DIR> has </CONST_DIR><LIMIT> 3000 </LIMIT><OPERATOR> LESS_OR_EQUAL </OPERATOR><CONST_TYPE> [LINEAR_CONSTRAINT] </CONST_TYPE> [is] <VAR> painkiller pill </VAR> [TIMES] <PARAM> 10 </PARAM><VAR> sleeping pill </VAR> [TIMES] <PARAM> 6 </PARAM></DECLARATION>
 \end{lstlisting}
 &
  \vspace{-2em}
 \begin{lstlisting}
<DECLARATION><CONST_DIR> has </CONST_DIR><LIMIT> 3000 </LIMIT><OPERATOR> LESS_OR_EQUAL </OPERATOR><CONST_TYPE> [LINEAR_CONSTRAINT] </CONST_TYPE> [is] <VAR> painkiller pill </VAR> [TIMES] <PARAM> 10 </PARAM><VAR> sleeping pill </VAR> [TIMES] <PARAM> 6 </PARAM></DECLARATION>
 \end{lstlisting}
 \\
 \hline
\end{tabular}
\caption{Comparison of predicted vs. gold IR for out-of-domain example (Health Science). In this example, our model predicted the wrong constraint type as lower bound instead of a ratio constraint. This shows that it is sometimes hard for the model to distinguish between similar constraint types. The rest of the declaration that follows the constraint type is also invalid.} \label{supp:ood-02}
\end{table}

\onecolumn
\begin{table}[ht]
\begin{tabular}{|p{7.5cm}|p{7.5cm}|}
 \hline
 \multicolumn{2}{|p{15cm}|}{A parent feeds their baby two flavors of baby food, apple and carrot, in order to meet the babies fat and folate requirements. Each serving of apple flavored baby food contains 2 units of fat and 5 units of folate. Each serving of carrot flavored baby food contains 4 units of fat and 3 units of folate. The baby does not like the carrot flavor, and therefore he must eat three times as many apple flavored baby food as carrot flavored baby food. However, he must eat at least 2 servings of carrot flavored baby food. If the baby can consume at most 100 units of folate, how many servings of each should he eat to maximize his fat intake?} \\
 \hline
  \multicolumn{1}{|c|}{Gold}&  \multicolumn{1}{|c|}{Pred} \\
 \hline
 \begin{lstlisting}
<DECLARATION><OBJ_DIR> maximize </OBJ_DIR><OBJ_NAME> fat intake </OBJ_NAME> [is] <VAR> apple flavored baby </VAR> [TIMES] <PARAM>(*@ \hlblue{2} @*)</PARAM>(*@ \hlblue{<VAR> carrot flavored baby </VAR> [TIMES] <PARAM> 4 </PARAM>}@*)</DECLARATION>
 \end{lstlisting}
 &
 \begin{lstlisting}
<DECLARATION><OBJ_DIR> maximize </OBJ_DIR><OBJ_NAME> fat intake </OBJ_NAME> [is] <VAR> apple flavored baby(*@ \hlred{food} @*)</VAR> [TIMES] <PARAM>(*@ \hlred{5} @*)</PARAM></DECLARATION>
 \end{lstlisting}
 \\
 \vspace{-2em}
 \begin{lstlisting}
<DECLARATION><CONST_DIR> must eat </CONST_DIR><OPERATOR> GREATER_OR_EQUAL </OPERATOR><CONST_TYPE> [XBY_CONSTRAINT] </CONST_TYPE><VAR> carrot flavored baby </VAR> [TIMES] <PARAM> three </PARAM> [is] <VAR> apple flavored baby food </VAR></DECLARATION>
 \end{lstlisting}
 &
 \vspace{-2em}
 \begin{lstlisting}
<DECLARATION><CONST_DIR> must eat </CONST_DIR><OPERATOR> GREATER_OR_EQUAL </OPERATOR><CONST_TYPE> [XBY_CONSTRAINT] </CONST_TYPE><VAR> carrot flavored baby (*@ \hlred{food} @*)</VAR> [TIMES] <PARAM> three </PARAM> [is] <VAR> apple flavored baby food </VAR></DECLARATION>
 \end{lstlisting}
 \\
  \vspace{-2em}
  \begin{lstlisting}
<DECLARATION><CONST_DIR> at least </CONST_DIR><OPERATOR> GREATER_OR_EQUAL </OPERATOR><LIMIT> 2 </LIMIT><CONST_TYPE> [LOWER_BOUND] </CONST_TYPE> [for] <VAR> carrot flavored baby </VAR></DECLARATION>
 \end{lstlisting}
 &
  \vspace{-2em}
 \begin{lstlisting}
<DECLARATION><CONST_DIR> at least </CONST_DIR><OPERATOR> GREATER_OR_EQUAL </OPERATOR><LIMIT> 2 </LIMIT><CONST_TYPE> [LOWER_BOUND] </CONST_TYPE> [for] <VAR> carrot flavored baby(*@ \hlred{food} @*)</VAR></DECLARATION>
 \end{lstlisting}
 \\
  \vspace{-2em}
  \begin{lstlisting}
<DECLARATION><CONST_DIR> at most </CONST_DIR><LIMIT> 100 </LIMIT><OPERATOR> LESS_OR_EQUAL </OPERATOR><CONST_TYPE> [LINEAR_CONSTRAINT] </CONST_TYPE> [is] <VAR> apple flavored baby </VAR> [TIMES] <PARAM>(*@ \hlblue{5} @*)</PARAM><VAR> carrot flavored baby </VAR> [TIMES] <PARAM>(*@ \hlblue{3} @*)</PARAM></DECLARATION>
 \end{lstlisting}
 &
  \vspace{-2em}
 \begin{lstlisting}
<DECLARATION><CONST_DIR> at most </CONST_DIR><LIMIT> 100 </LIMIT><OPERATOR> LESS_OR_EQUAL </OPERATOR><CONST_TYPE> [LINEAR_CONSTRAINT] </CONST_TYPE> [is] <VAR> apple flavored baby(*@ \hlred{food} @*)</VAR> [TIMES] <PARAM>(*@ \hlred{2} @*) </PARAM><VAR> carrot flavored baby(*@ \hlred{food} @*)</VAR> [TIMES] <PARAM>(*@ \hlred{4} @*)</PARAM></DECLARATION>
 \end{lstlisting}
 \\
 \hline
\end{tabular}
\caption{Comparison of predicted vs. gold IR for out-of-domain example (Health Science). In this example, the model produced an erroneous declaration in the objective and the last constraint. The wrong data parameters, which should describe the objective function, are instead copied into the last constraint. This example illustrates the difficulty of parsing an input document that is ambiguous.}\label{supp:ood-03}
\end{table}

\onecolumn
\begin{table}[ht]
\begin{tabular}{|p{7.5cm}|p{7.5cm}|}
 \hline
 \multicolumn{2}{|p{15cm}|}{A chocolate company can transport their boxes of chocolate either using their own vans or by renting trucks. Their vans can transport 50 boxes per trip while a truck can transport 80 boxes per trip. Since they own their vans, the cost per van trip is \$30 while the cost per truck trip is \$50. The company needs to transport at least 1500 boxes of chocolate and they have a budget of \$1000. Since the vans also provide advertising, the number of trips by van must be larger than the number of trips by trucks. How many of trip by each should be done to minimize the total number of trips?} \\
 \hline
  \multicolumn{1}{|c|}{Gold}&  \multicolumn{1}{|c|}{Pred} \\
 \hline
 \begin{lstlisting}
<DECLARATION><OBJ_DIR> minimize </OBJ_DIR><OBJ_NAME> number of trips </OBJ_NAME> [is] <VAR> van(*@\hlblue{s}@*) </VAR> [TIMES] <PARAM> ONE </PARAM><VAR> truck(*@\hlblue{s}@*) </VAR> [TIMES] <PARAM> ONE </PARAM></DECLARATION>
 \end{lstlisting}
 &
 \begin{lstlisting}
<DECLARATION><OBJ_DIR> minimize </OBJ_DIR><OBJ_NAME> (*@\hlred{total}@*) number of trips </OBJ_NAME> [is] <VAR> van </VAR> [TIMES] <PARAM> ONE </PARAM><VAR> truck </VAR> [TIMES] <PARAM> ONE </PARAM></DECLARATION>
 \end{lstlisting}
 \\
 \vspace{-2em}
 \begin{lstlisting}
<DECLARATION><CONST_DIR> at least </CONST_DIR><LIMIT> 1500 </LIMIT><OPERATOR> GREATER_OR_EQUAL </OPERATOR><CONST_TYPE> [LINEAR_CONSTRAINT] </CONST_TYPE> [is] <VAR> vans </VAR> [TIMES] <PARAM> 50 </PARAM><VAR> truck </VAR> [TIMES] <PARAM> 80 </PARAM></DECLARATION>
 \end{lstlisting}
 &
 \vspace{-2em}
 \begin{lstlisting}
<DECLARATION><CONST_DIR> at least </CONST_DIR><LIMIT> 1500 </LIMIT><OPERATOR> GREATER_OR_EQUAL </OPERATOR><CONST_TYPE> [LINEAR_CONSTRAINT] </CONST_TYPE> [is] <VAR> vans </VAR> [TIMES] <PARAM> 50 </PARAM><VAR> truck(*@\hlred{s}@*) </VAR> [TIMES] <PARAM> 80 </PARAM></DECLARATION>
 \end{lstlisting}
 \\
  \vspace{-2em}
  \begin{lstlisting}
<DECLARATION><CONST_DIR> budget </CONST_DIR><LIMIT> 1000 </LIMIT><OPERATOR> LESS_OR_EQUAL </OPERATOR><CONST_TYPE> [LINEAR_CONSTRAINT] </CONST_TYPE> [is] <VAR> van </VAR> [TIMES] <PARAM> 30 </PARAM><VAR> truck </VAR> [TIMES] <PARAM> 50 </PARAM></DECLARATION>
 \end{lstlisting}
 &
  \vspace{-2em}
 \begin{lstlisting}
<DECLARATION><CONST_DIR> budget </CONST_DIR><LIMIT> 1000 </LIMIT><OPERATOR> LESS_OR_EQUAL </OPERATOR><CONST_TYPE> [LINEAR_CONSTRAINT] </CONST_TYPE> [is] <VAR> van </VAR> [TIMES] <PARAM> 30 </PARAM><VAR> truck </VAR> [TIMES] <PARAM> 50 </PARAM></DECLARATION>
 \end{lstlisting}
 \\
  \vspace{-2em}
  \begin{lstlisting}
<DECLARATION><CONST_DIR> must be larger than </CONST_DIR><OPERATOR> GREATER_OR_EQUAL </OPERATOR><CONST_TYPE>(*@ \hlblue{[XY\_CONSTRAINT]} @*)</CONST_TYPE>(*@ \hlblue{<VAR> trucks </VAR> [is] <VAR> van </VAR>} *@)</DECLARATION>
 \end{lstlisting}
 &
  \vspace{-2em}
 \begin{lstlisting}
<DECLARATION><CONST_DIR> must be larger than </CONST_DIR><LIMIT> 500 </LIMIT><OPERATOR> GREATER_OR_EQUAL </OPERATOR><CONST_TYPE>(*@  \hlred{[LINEAR\_CONSTRAINT]} @*)</CONST_TYPE>(*@  \hlred{[is] <VAR> vans </VAR> [TIMES] <PARAM> 50 </PARAM><VAR> trucks </VAR> [TIMES] <PARAM> 80 </PARAM>} @*)</DECLARATION>
 \end{lstlisting}
 \\
 \hline
\end{tabular}
\caption{Comparison of predicted vs. gold IR for out-of-domain example (Transportation example). In this example, our model detects a linear constraint instead of a balance constraint. In fact, the balance constraints are less frequent in the training dataset whereas the linear constraints are the majority ones. This can explain this type of error as the constraint types are imbalanced in the training dataset.}\label{supp:ood-04}
\end{table}

\onecolumn
\begin{table}[ht]
\begin{tabular}{|p{7.5cm}|p{7.5cm}|}
 \hline
 \multicolumn{2}{|p{15cm}|}{A toy store decides to deliver gifts using two shipping companies, a new one and an old one. The new company can deliver 50 gifts per trip while the old company can deliver 70 gifts per trip. The new company uses 30 liters of diesel per trip while the old company uses 40 liters of diesel per trip. The toy store needs to deliver at least 1000 gifts. There can be at most 15 trips made by the new company. In order to make sure that the old company does not go out of business, at least 40\% of all trips must be made by the old company. How many trips should each company make to minimize the total amount of diesel used?} \\
 \hline
  \multicolumn{1}{|c|}{Gold}&  \multicolumn{1}{|c|}{Pred} \\
 \hline
 \begin{lstlisting}
<DECLARATION><OBJ_DIR> minimize </OBJ_DIR><OBJ_NAME> total amount of diesel </OBJ_NAME> [is] <VAR> new company </VAR> [TIMES] <PARAM> 30 </PARAM><VAR> old company </VAR> [TIMES] <PARAM> 40 </PARAM></DECLARATION>
 \end{lstlisting}
 &
 \begin{lstlisting}
<DECLARATION><OBJ_DIR> minimize </OBJ_DIR><OBJ_NAME> total amount of diesel </OBJ_NAME> [is] <VAR> new company </VAR> [TIMES] <PARAM> 30 </PARAM><VAR> old company </VAR> [TIMES] <PARAM> 40 </PARAM></DECLARATION>
 \end{lstlisting}
 \\
 \vspace{-2em}
 \begin{lstlisting}
<DECLARATION><CONST_DIR> at least </CONST_DIR><LIMIT> 1000 </LIMIT><OPERATOR> GREATER_OR_EQUAL </OPERATOR><CONST_TYPE> [LINEAR_CONSTRAINT] </CONST_TYPE> [is] <VAR> new company </VAR> [TIMES] <PARAM> 50 </PARAM><VAR> old company </VAR> [TIMES] <PARAM> 70 </PARAM></DECLARATION>
 \end{lstlisting}
 &
 \vspace{-2em}
 \begin{lstlisting}
<DECLARATION><CONST_DIR> at least </CONST_DIR><LIMIT> 1000 </LIMIT><OPERATOR> GREATER_OR_EQUAL </OPERATOR><CONST_TYPE> [LINEAR_CONSTRAINT] </CONST_TYPE> [is] <VAR> new company </VAR> [TIMES] <PARAM> 50 </PARAM><VAR> old company </VAR> [TIMES] <PARAM> 70 </PARAM></DECLARATION>
 \end{lstlisting}
 \\
  \vspace{-2em}
  \begin{lstlisting}
<DECLARATION><CONST_DIR> at most </CONST_DIR><LIMIT> 15 </LIMIT><OPERATOR> LESS_OR_EQUAL </OPERATOR><CONST_TYPE>(*@ \hlblue{[UPPER\_BOUND]} @*)</CONST_TYPE>(*@ \hlblue{[for] <VAR> new company </VAR>}@*)</DECLARATION>
 \end{lstlisting}
 &
  \vspace{-2em}
 \begin{lstlisting}
<DECLARATION><CONST_DIR> at most </CONST_DIR><LIMIT> 15 </LIMIT><OPERATOR> LESS_OR_EQUAL </OPERATOR><CONST_TYPE>(*@ \hlred{[LINEAR\_CONSTRAINT]} @*)</CONST_TYPE>(*@ \hlred{[is] <VAR> new company </VAR> [TIMES] <PARAM> 50 </PARAM><VAR> old company </VAR> [TIMES] <PARAM> 70 </PARAM>}@*)</DECLARATION>
 \end{lstlisting}
 \\
  \vspace{-2em}
  \begin{lstlisting}
<DECLARATION><CONST_DIR> at least </CONST_DIR><LIMIT>(*@ \hlblue{40\%} @*)</LIMIT><OPERATOR> GREATER_OR_EQUAL </OPERATOR><CONST_TYPE>(*@ \hlblue{[RATIO\_CONSTRAINT]} @*)</CONST_TYPE>(*@ \hlblue{[for] <VAR> old company </VAR>}@*)</DECLARATION>
 \end{lstlisting}
 &
  \vspace{-2em}
 \begin{lstlisting}
<DECLARATION><CONST_DIR> at least </CONST_DIR><LIMIT>(*@ \hlred{1000} @*)</LIMIT><OPERATOR> GREATER_OR_EQUAL </OPERATOR><CONST_TYPE>(*@ \hlred{[LINEAR\_CONSTRAINT]} @*)</CONST_TYPE>(*@ \hlred{[is] <VAR> new company </VAR> [TIMES] <PARAM> 50 </PARAM><VAR> old company </VAR> [TIMES] <PARAM> 70 </PARAM>}@*)</DECLARATION>
 \end{lstlisting}
 \\
 \hline
\end{tabular}
\caption{Comparison of predicted vs. gold IR for out-of-domain example (Transportation example). In this example, the model incorrectly generated the expressions for the last two constraints. In fact, it detects the wrong constraint types and produced the same invalid algebraic expression. This suggests that the generation could be made more precise by adding additional constrain context into the declaration prompt to distinguish between different constraints.}\label{supp:ood-06}
\end{table}

\onecolumn
\begin{table}[ht]
\begin{tabular}{|p{7.5cm}|p{7.5cm}|}
 \hline
 \multicolumn{2}{|p{15cm}|}{A zoo needs to transport their monkeys to the vet either by bus or by car. A bus can transport 20 monkeys per trip and takes 30 minutes. A car can transport 6 monkeys per trip and takes 15 minutes. There can be at most 10 bus trips. In addition, since the monkeys get aggressive when there are too many in one place at least 60\% of the trips should be by car. If the zoo needs to transport 300 monkeys, how many trips of each should be done to minimize the total time required to transport the monkeys?} \\
 \hline
  \multicolumn{1}{|c|}{Gold}&  \multicolumn{1}{|c|}{Pred} \\
 \hline
 \begin{lstlisting}
<DECLARATION><OBJ_DIR> minimize </OBJ_DIR><OBJ_NAME> total time </OBJ_NAME> [is] <VAR>(*@ \hlblue{bus} @*)</VAR> [TIMES] <PARAM> 30 </PARAM><VAR> car </VAR> [TIMES] <PARAM> 15 </PARAM></DECLARATION>
 \end{lstlisting}
 &
 \begin{lstlisting}
<DECLARATION><OBJ_DIR> minimize </OBJ_DIR><OBJ_NAME> total time </OBJ_NAME> [is] <VAR>(*@ \hlred{monkeys} @*)</VAR> [TIMES] <PARAM> 30 </PARAM><VAR> car(*@\hlred{s}@*) </VAR> [TIMES] <PARAM> 15 </PARAM></DECLARATION>
 \end{lstlisting}
 \\
 \vspace{-2em}
 \begin{lstlisting}
<DECLARATION><CONST_DIR> at most </CONST_DIR><LIMIT> 10 </LIMIT><OPERATOR> LESS_OR_EQUAL </OPERATOR><CONST_TYPE> [UPPER_BOUND] </CONST_TYPE> [for] <VAR> bus </VAR></DECLARATION>
 \end{lstlisting}
 &
 \vspace{-2em}
 \begin{lstlisting}
<DECLARATION><CONST_DIR> at most </CONST_DIR><LIMIT> 10 </LIMIT><OPERATOR> LESS_OR_EQUAL </OPERATOR><CONST_TYPE> [UPPER_BOUND] </CONST_TYPE> [for] <VAR> bus </VAR></DECLARATION>
 \end{lstlisting}
 \\
  \vspace{-2em}
  \begin{lstlisting}
<DECLARATION><CONST_DIR> at least </CONST_DIR><LIMIT> 60% </LIMIT><OPERATOR> GREATER_OR_EQUAL </OPERATOR><CONST_TYPE> [RATIO_CONSTRAINT] </CONST_TYPE> [for] <VAR> car </VAR></DECLARATION>
 \end{lstlisting}
 &
  \vspace{-2em}
 \begin{lstlisting}
<DECLARATION><CONST_DIR> at least </CONST_DIR><LIMIT> 60% </LIMIT><OPERATOR> GREATER_OR_EQUAL </OPERATOR><CONST_TYPE> [RATIO_CONSTRAINT] </CONST_TYPE> [for] <VAR> car </VAR></DECLARATION>
 \end{lstlisting}
 \\
  \vspace{-2em}
  \begin{lstlisting}
<DECLARATION><CONST_DIR> needs </CONST_DIR><LIMIT> 300 </LIMIT><OPERATOR> GREATER_OR_EQUAL </OPERATOR><CONST_TYPE> [LINEAR_CONSTRAINT] </CONST_TYPE> [is] <VAR> bus </VAR> [TIMES] <PARAM> 20 </PARAM><VAR> car </VAR> [TIMES] <PARAM> 6 </PARAM></DECLARATION>
 \end{lstlisting}
 &
  \vspace{-2em}
 \begin{lstlisting}
<DECLARATION><CONST_DIR> needs </CONST_DIR><LIMIT> 300 </LIMIT><OPERATOR> GREATER_OR_EQUAL </OPERATOR><CONST_TYPE> [LINEAR_CONSTRAINT] </CONST_TYPE> [is] <VAR> bus </VAR> [TIMES] <PARAM> 20 </PARAM><VAR> car </VAR> [TIMES] <PARAM> 6 </PARAM></DECLARATION>
 \end{lstlisting}
 \\
 \hline
\end{tabular}
\caption{Comparison of predicted vs. gold IR for out-of-domain example (Transportation example). In this example, the model detects an invalid decision variable "monkeys" in the predicted objective declaration.} \label{supp:ood-7}
\end{table}

\begin{table*}[ht]
\centering
\fontsize{10}{12}\selectfont
\resizebox{\textwidth}{!}{\begin{tabular}{p{2cm}llllllllll}
\toprule
\multicolumn{1}{c}{ } & \multicolumn{1}{c}{ } & \multicolumn{3}{c}{Rouge-1} & \multicolumn{3}{c}{Rouge-2}  & \multicolumn{3}{c}{Rouge-L}\\
\cmidrule(l{3pt}r{3pt}){3-5} \cmidrule(l{3pt}r{3pt}){6-8} \cmidrule(l{3pt}r{3pt}){9-11} 
Method & Accuracy & Recall & Precision & F1 & Recall & Precision & F1  & Recall & Precision & F1  \\
\midrule
OptGen & 0.61 & 0.89  & 0.89 & 0.89 & 0.80 & 0.80 & 0.79 & 0.86 & 0.86 & 0.86 \\

\bottomrule
\end{tabular}}
\vspace{-6pt}
\caption{\label{appendix:tab:dev} Performance of our model on development set. Rouge scores included declaration tags to illustrate how well the model is able to reproduce syntax.}
\end{table*}



\end{document}